\documentclass[10pt,twocolumn,letterpaper]{article}

\usepackage{cvpr}              

\usepackage{algorithm}
\usepackage{algpseudocode}
\usepackage{booktabs, tabularx, colortbl, xcolor, multirow}
\usepackage{pifont}
\newcommand{\cmark}{\ding{51}} 
\newcommand{\xmark}{\ding{55}} 
\usepackage[most]{tcolorbox}
\usepackage{verbatim}

\definecolor{cvprblue}{rgb}{0.21,0.49,0.74}
\usepackage[pagebackref,breaklinks,colorlinks,allcolors=cvprblue]{hyperref}

\def\paperID{} 
\def\confName{CVPR}
\def\confYear{2026}

\title{VisRefiner: Learning from Visual Differences for Screenshot-to-Code Generation}

\author{
Jie Deng$^{1,2}$ \quad 
Kaichun Yao$^{1}$ \quad 
Libo Zhang$^{1}$\thanks{Corresponding author}\\
$^{1}$Institute of Software, Chinese Academy of Sciences\\
$^{2}$University of Chinese Academy of Sciences\\
{\tt\small dengjie242@mails.ucas.ac.cn \quad libo@iscas.ac.cn}
}

\begin{document}
\maketitle
\begin{abstract}
Screenshot-to-code generation aims to translate user interface screenshots into executable frontend code that faithfully reproduces the target layout and style. Existing multimodal large language models perform this mapping directly from screenshots but are trained without observing the visual outcomes of their generated code. In contrast, human developers iteratively render their implementation, compare it with the design, and learn how visual differences relate to code changes. Inspired by this process, we propose \textbf{VisRefiner}, a training framework that enables models to learn from visual differences between rendered predictions and reference designs. We construct difference-aligned supervision that associates visual discrepancies with corresponding code edits, allowing the model to understand how appearance variations arise from implementation changes. Building on this, we introduce a reinforcement learning stage for self-refinement, where the model improves its generated code by observing both the rendered output and the target design, identifying their visual differences, and updating the code accordingly. Experiments show that VisRefiner substantially improves single-step generation quality and layout fidelity, while also endowing models with strong self-refinement ability. These results demonstrate the effectiveness of learning from visual differences for advancing screenshot-to-code generation.
\end{abstract}
    
\section{Introduction}
\label{sec:intro}

\begin{figure}[t]
    \centering
    \includegraphics[width=\linewidth]{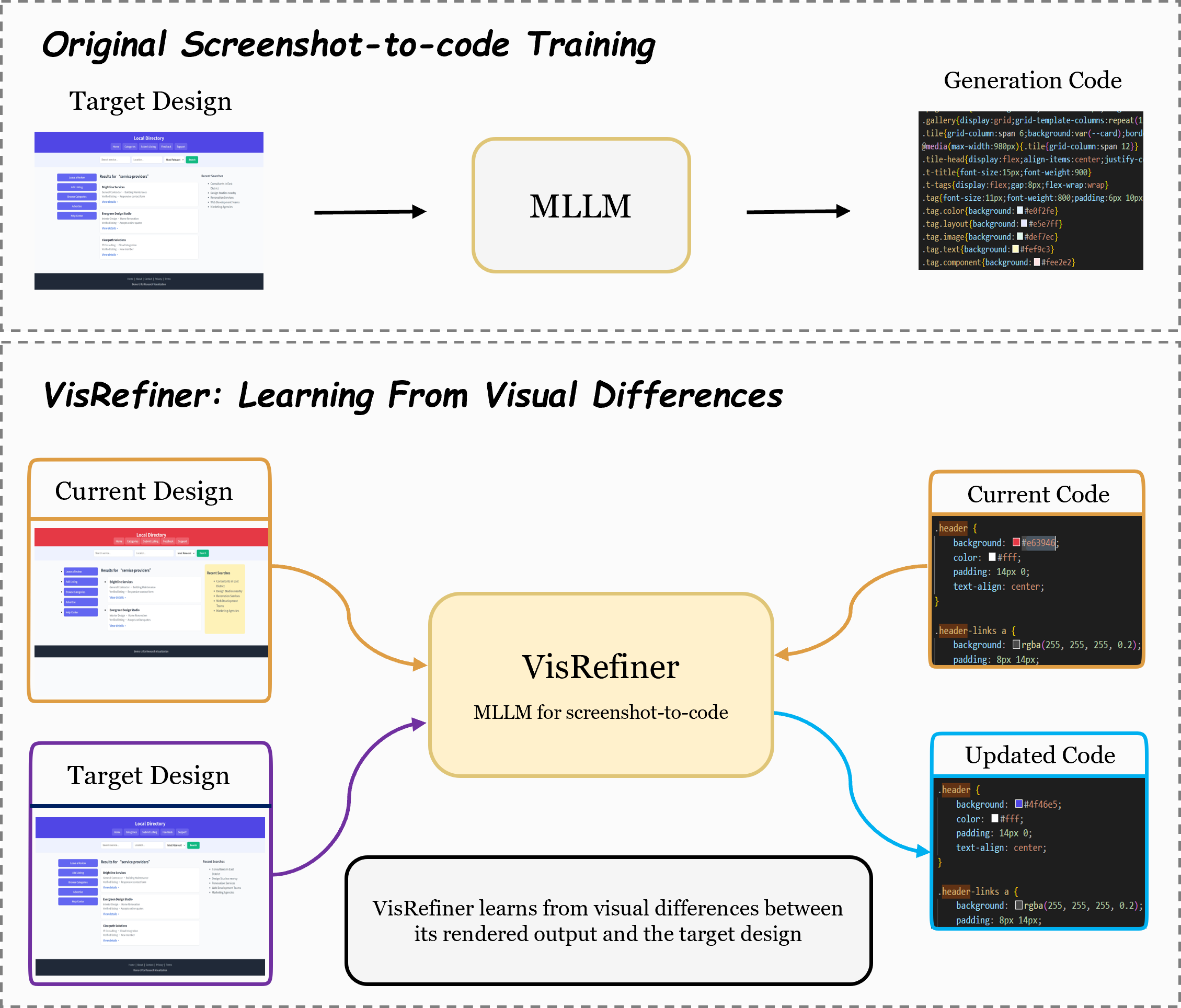}
    \caption{
    Comparison of training paradigms for screenshot-to-code generation.
    (a) Original approaches train models through one-way supervised mapping from target designs to code.  
    (b) \textbf{VisRefiner} introduces a difference-driven training paradigm, where the model learns from visual discrepancies between its rendered output and the target design.
    }
    \label{fig:training_comparison}
\end{figure}

Automatically translating user interface (UI) screenshots into executable frontend code has become an emerging topic in multimodal learning. The goal is to convert a static visual representation of a webpage into structured HTML and CSS that can faithfully reproduce its layout and style. Automating this process reduces repetitive implementation effort, accelerates prototyping, and lowers the barrier for non-experts to create functional web interfaces.

Early work addressed this task using rule-based heuristics or convolutional architectures that convert visual layouts into structured HTML or token sequences~\cite{beltramelli2018pix2code,acsirouglu2019automatic,robinson2019sketch2code,xu2021image2emmet}. Subsequent research explored layout synthesis from schematic or wireframe-style inputs~\cite{zhao2021guigan} and leveraged pretrained language models to generate code from structured prompts or layout metadata~\cite{wang2021codet5,wang2023codet5+}. With the emergence of multimodal large language models (MLLMs), recent approaches have enabled end-to-end screenshot-to-code generation without intermediate supervision~\cite{laurenccon2024unlocking,si2024design2code,zhang2024internlm,achiam2023gpt,anthropic2024claude3}. These models demonstrate strong capability in visual understanding and structured code synthesis, providing a promising foundation for automatic UI generation.

To further improve generation quality, recent studies have explored strategies that enhance model reasoning or introduce refinement after the initial output. Some methods decompose the interface into regions, components, or hierarchical layouts to simplify generation and improve structural alignment~\cite{wan2025divide,gui2025uicopilot,chen2025designcoder,xiao2024prototype2code}. Others enrich model input with intermediate structural cues such as component trees or semantic grouping to guide layout organization~\cite{xiao2024ui,zhou2025declarui}. Complementary approaches perform post-generation refinement by comparing rendered results with target designs. These include prompting models to revise code based on visual discrepancies~\cite{si2024design2code,chen2025designcoder}, selecting high-quality outputs through similarity-based filtering~\cite{wu2024uicoder}, or applying rule-based corrections derived from layout differences~\cite{zhou2024bridging}. While these methods have improved layout fidelity and structural coherence, they generally treat visual differences as external diagnostic signals rather than internal learning cues.

However, effective UI development often depends on the ability to interpret and learn from visual differences. Human developers iteratively render a webpage, compare it to the target design, identify visual inconsistencies, and refine the underlying code. This render–compare–revise process provides rich and localized difference signals that link appearance to implementation. Current MLLM-based systems lack this mechanism.

We propose \textbf{VisRefiner}, a training framework that enables multimodal large language models to learn directly from visual differences between rendered outputs and target screenshots.  
Rather than treating discrepancies as post-hoc feedback, VisRefiner integrates difference signals into training, allowing the model to link appearance changes to corresponding code edits.
The framework consists of two stages.  
First, \emph{difference-aligned supervision} connects localized visual changes with corrective code updates, grounding the model’s understanding of how implementation affects appearance.  
Second, \emph{reinforcement learning with self-refinement} allows the model to generate, compare, and improve its code through visual rewards that reflect reduction in discrepancy.  
This process encourages iterative reasoning and internalizes refinement behavior.

Experiments show that VisRefiner improves visual fidelity and layout consistency across multiple benchmarks, while maintaining stable self-refinement capability.  
By learning from visual differences, VisRefiner shifts screenshot-to-code generation from a feed-forward prediction task to a difference-driven learning paradigm, advancing multimodal large language models toward more human-like reasoning in structured code synthesis.

\textbf{Contributions.}
\begin{itemize}
    \item We present \textbf{VisRefiner}, a framework that enables multimodal large language models to learn from visual differences between rendered outputs and target designs, systematically applying difference-driven learning to screenshot-to-code generation.
    \item We develop \textbf{VisDiffUI}, a dataset and data pipeline that align visual differences with corresponding code edits, providing a resource for studying appearance-to-implementation correspondences.
    \item We conduct extensive experiments demonstrating that VisRefiner achieves consistent improvements in visual fidelity, layout alignment, and self-improvement capability across multiple screenshot-to-code benchmarks.
\end{itemize}

\begin{figure*}[t]
    \centering
    \includegraphics[width=\textwidth]{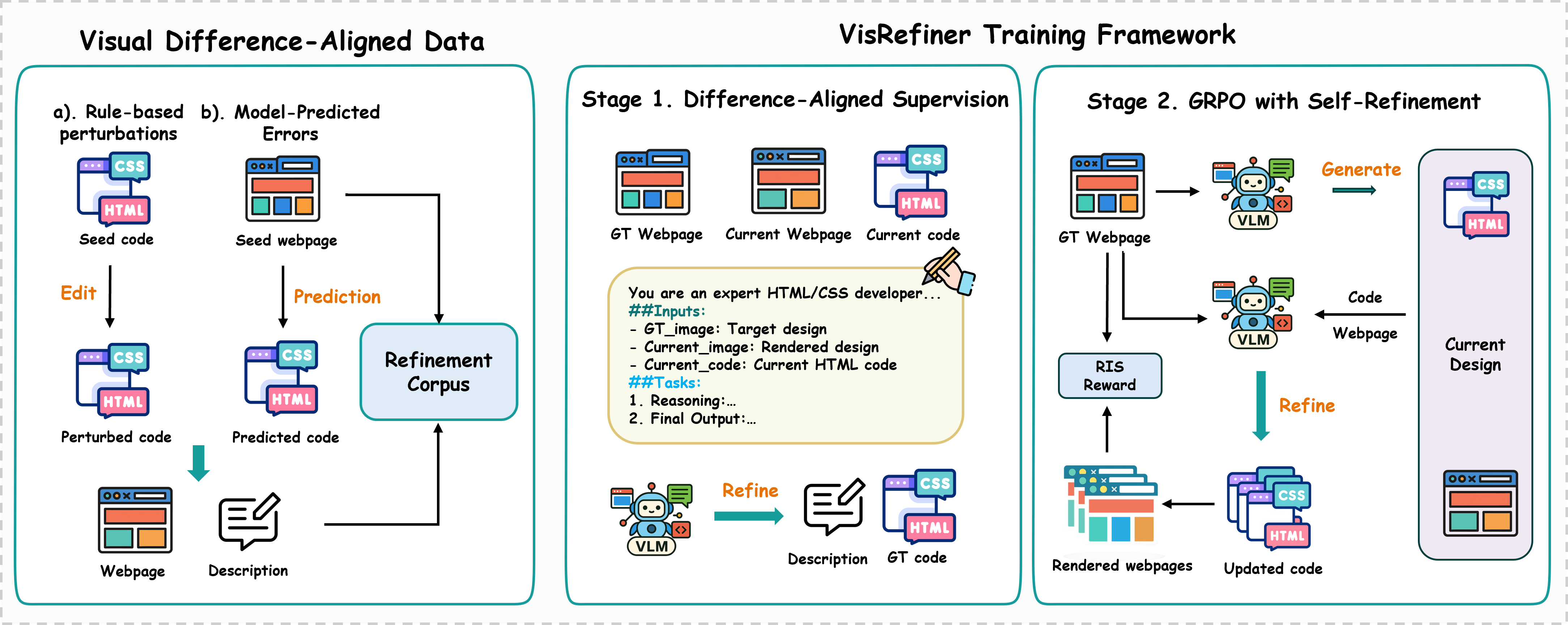}
    \caption{
    Overview of the proposed \textbf{VisRefiner} framework. 
    The process begins with constructing a visual difference-aligned corpus that provides paired examples of visual deviations and their corresponding code edits. 
    \textbf{Stage~1} learns to interpret these visual differences through difference-aligned supervision, grounding visual understanding in code updates. 
    \textbf{Stage~2} applies GRPO-based optimization with self-refinement, where the model refines its own predictions based on perceptual rewards derived from rendered similarity improvements. 
    Together, these stages enable multimodal LLMs to learn directly from visual differences during training.
    }
    \label{fig:overview}
\end{figure*}

\section{Related Work}
\label{sec:relatedwork}

\subsection{Screenshot-to-Code Generation}
Screenshot-to-code generation aims to translate visual UI designs into structured frontend code. Early work used rule-based or convolutional models to map mockups to HTML or token sequences~\cite{beltramelli2018pix2code,acsirouglu2019automatic,robinson2019sketch2code,xu2021image2emmet}, followed by studies on layout synthesis from wireframes~\cite{zhao2021guigan} and code generation from structured prompts or design metadata with pretrained language models~\cite{wang2021codet5,wang2023codet5+}. To enhance layout fidelity, later methods incorporated visual reasoning modules that extract component hierarchies or structural representations to guide generation~\cite{xiao2024prototype2code,zhou2025declarui,lee2023pix2struct,hong2024cogagent,xiao2024ui}. 
With Multimodal Large Language Models (MLLMs)~\cite{achiam2023gpt,anthropic2024claude3}, end-to-end systems have emerged~\cite{laurenccon2024unlocking,si2024design2code,zhang2024internlm}, achieving fluent code generation yet often lacking precise visual alignment. Recent work thus focuses on structural and stylistic accuracy via improved reasoning or refinement: some decompose interfaces into hierarchical components to simplify generation~\cite{wan2025divide,gui2025uicopilot,chen2025designcoder}, while others apply post-generation refinement through visual comparison~\cite{si2024design2code,chen2025designcoder,wu2024uicoder,zhou2024bridging}. Despite reducing certain errors, these approaches depend on explicit decomposition or heuristic correction and do not enable learning from visual differences.

\subsection{Multimodal Large Language Models}

Large Language Models (LLMs) have rapidly advanced natural language understanding and generation, demonstrating strong capability across diverse domains. Code generation has become a particularly successful application, with numerous models achieving high proficiency in translating natural language or structured prompts into executable programs~\cite{nijkamp2022codegen,guo2023longcoder,shojaee2023execution,roziere2023code,guo2024deepseek}. Multimodal Large Language Models(MLLMs) extend LLMs by integrating visual perception, thereby unifying vision and language understanding within a single generative framework~\cite{liu2023visual,lu2024deepseek,hong2024cogagent,zhu2023minigpt,chen2024internvl, tan2025ocr, qiao2025v}.  These models have shown strong generalization across a variety of vision–language tasks. Recently, an emerging direction focuses on \textit{multimodal code generation}, where MLLMs translate visual inputs into structured and executable representations~\cite{khan2024self,zhao2025vincicoder}. Existing works span multiple domains such as chart and plot generation~\cite{wu2025plot2code,zhao2025chartcoder}, vector graphic and SVG synthesis~\cite{rodriguez2025starvector,rodriguez2025rendering}, web or UI code generation and so on.
\section{Preliminary}
\label{sec:prelim}

\subsection{Task Definition}

\paragraph{Screenshot-to-Code.}
Screenshot-to-code aims to learn a mapping from a target user interface (UI) screenshot $I_{gt}$ to structured frontend code $C_{gen}$ (e.g., HTML/CSS), where rendering $C_{gen}$ visually reproduces the input design:
\begin{equation}
C_{gen} = f_{\text{gen}}(I_{gt}), \quad 
\mathrm{Render}(C_{gen}) \approx I_{gt}.
\label{eq:s2c}
\end{equation}
This task requires the model to reason over visual layout, component hierarchy, and style in order to generate code that faithfully reconstructs the appearance of $I_{gt}$.

\paragraph{Visual Refinement.}
To enable the model to learn from its own outputs, we define a refinement task where the model observes both the rendered result of its prediction and the target screenshot.  
Given the current code $C_t$, its rendered image $I_t = \mathrm{Render}(C_t)$, and the ground-truth target $I_{gt}$, the model generates updated code $C_{t+1}$:
\begin{equation}
C_{t+1} = f_{\text{refine}}(I_t, I_{gt}, C_t),
\label{eq:refine}
\end{equation}
such that the new rendering $\mathrm{Render}(C_{t+1})$ becomes visually closer to $I_{gt}$ than $\mathrm{Render}(C_t)$.  
Although the refinement process can, in principle, be applied iteratively, we formulate training as a collection of single-step refinement instances.  
Each instance teaches the model to identify visual discrepancies and generate code updates that reduce them, allowing refinement behavior to emerge naturally during inference.

\begin{figure*}[t]
    \centering
    \includegraphics[width=\textwidth]{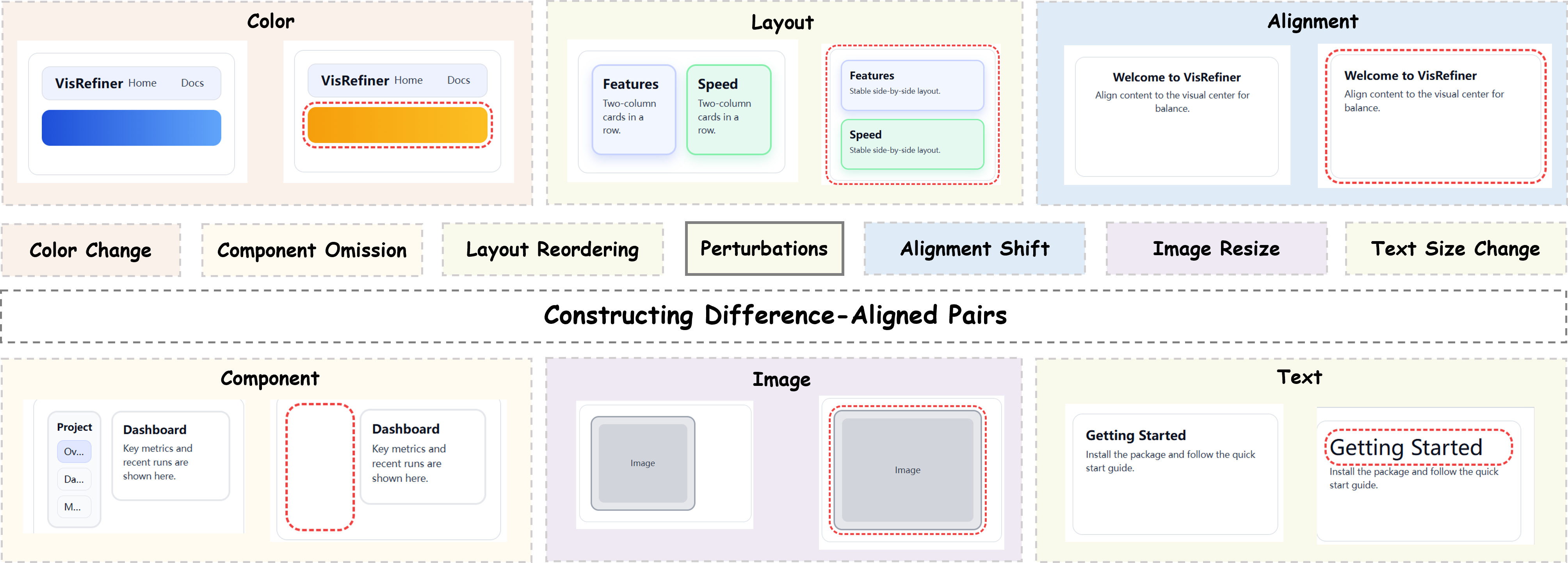}
    \caption{
    Representative categories of \textbf{difference-aligned perturbations} used in constructing training pairs.  
    Each group shows a reference UI on the left and its perturbed counterpart on the right.  
    The perturbations cover six visual dimensions including color, layout, alignment, component, image, and text.  
    They introduce localized inconsistencies such as color drift, misalignment, component removal, and text resizing, which establish fine-grained supervision linking code edits to perceptual differences.
    }
    \label{fig:perturb_pipeline}
\end{figure*}

\section{VisRefiner}
\label{sec:method}

\subsection{Overview}
\label{sec:overview}

We propose \textbf{VisRefiner}, a difference-driven visual learning framework that enables multimodal large language models to improve code generation by learning from visual discrepancies between rendered results and target designs.  
Unlike previous screenshot-to-code systems that rely purely on forward supervision, VisRefiner integrates visual feedback into training, encouraging the model to internalize a refinement-oriented process similar to human developers.

As illustrated in Figure~\ref{fig:overview}, the framework consists of two stages:  
\textbf{(1) Difference-Aligned Supervision.} The model learns to associate localized visual differences with corresponding code edits, grounding implementation changes in perceptual cues.  
\textbf{(2) Reinforcement Learning with Self-Refinement.} Building on this foundation, the model further refines its outputs through group-relative policy optimization (GRPO), guided by perceptual rewards that quantify visual improvement.

Together, these stages integrate self-refinement into the training process, enabling the model to understand how code modifications affect visual outcomes and to produce more accurate and consistent UI implementations.

\subsection{Difference-Aligned Supervision}
\label{sec:sft_stage}

The first stage, \textbf{Difference-Aligned Supervision}, establishes an explicit correspondence between observed visual discrepancies and the code updates required to resolve them.  
We construct a corpus of \textit{difference-aligned refinement pairs}, where each pair captures two UI states, an imperfect implementation and its corrected counterpart, together with their associated code.  
Formally, each training instance is represented as $(I_t, I_{gt}, C_t, C_{t+1})$, where $C_t$ is a perturbed or model-predicted implementation, $I_t = \mathrm{Render}(C_t)$ is its rendered image, $I_{gt}$ is the target screenshot, and $C_{t+1}$ denotes the refined code.  
During training, the model receives $(I_t, I_{gt}, C_t)$ as input and predicts $C_{t+1}$ by minimizing the negative log-likelihood:
\begin{equation}
\mathcal{L}_{\text{SFT}} = - \log P_\theta(C_{t+1} \mid I_t, I_{gt}, C_t),
\label{eq:sft}
\end{equation}
where the target code is generated autoregressively, conditioned on both the visual comparison between $I_t$ and $I_{gt}$ and the contextual semantics of $C_t$.

\paragraph{Constructing Difference-Aligned Pairs.}
To provide diverse and informative supervision, we construct refinement pairs from two complementary sources: rule-based perturbations and model-predicted imperfections.

\textbf{(1) Rule-based perturbations.}  
We design a library of visually grounded perturbation rules that simulate typical UI inconsistencies encountered during development. As shown in Figure \ref{fig:perturb_pipeline}, each rule modifies the underlying HTML or CSS tokens in a controlled way, generating visually perceivable yet structurally localized deviations.  
The perturbations cover six major categories: \textit{color}, \textit{layout}, \textit{alignment}, \textit{component}, \textit{image}, and \textit{text}.  

Specifically,  
\textbf{color} perturbations alter palette tokens or gradients, producing hue shifts or inconsistent theme bindings;  
\textbf{layout} rearranges grid or flex structures, resulting in misplaced or overlapping elements;  
\textbf{alignment} introduces margin or offset shifts that disturb visual balance;  
\textbf{component} removes or substitutes interface elements such as sidebars, buttons, or footers;  
\textbf{image} adjusts aspect ratios or removes placeholder to simulate missing or resized assets;  
and \textbf{text} changes font size, weight, or truncates content to mimic incomplete textual rendering.  

Multiple perturbations can be composed on a single sample, creating compound discrepancies that better reflect real-world interface variations.  
Because each transformation is rule-defined, the mapping between modified visual regions and the corresponding code fragments remains precisely traceable.  
The perturbed code $C_t$ and render $I_t$ serve as the input state, while the original implementation $(C_{t+1}, I_{gt})$ provides the supervision target.

\textbf{(2) Model-predicted imperfections.}  
To complement the synthetic perturbations, we also incorporate imperfect outputs produced by an earlier-stage model.  
These samples naturally contain realistic artifacts such as misplaced containers, redundant wrappers, missing tokens, or incorrect color bindings, which are difficult to reproduce by rules alone.  
Each predicted pair $(I_t, C_t)$ is combined with its corresponding ground truth $(I_{gt}, C_{t+1})$, forming a richer and more diverse distribution of visual discrepancies.  
This hybrid construction enables the dataset to capture both structured and free-form variation patterns, thereby promoting a more generalizable understanding of visual differences.

\subsection{Reinforcement Learning with Self-Refinement}
\label{sec:rl_stage}

Building on the difference-aligned supervision stage, we introduce a reinforcement optimization framework that enables the model to refine its outputs through visual feedback.  
In this stage, visual discrepancies between rendered predictions and target designs are quantified as continuous learning signals.  
The model renders its code, compares the result with the reference design, and updates its policy according to the degree of visual improvement.  
Through this process, it learns self-refinement behavior and progressively enhances generation quality based on visual evidence.

\paragraph{Quantifying Visual Improvement.}
We define a \textbf{Relative Improvement Score (RIS)} that evaluates how effectively the model reduces the visual difference between its rendered output and the reference design. Let $s_t = \operatorname{CLIP}(I_t, I_{gt})$ and $s_{t+1} = \operatorname{CLIP}(I_{t+1}, I_{gt})$ denote the CLIP-based perceptual similarities before and after refinement. 
\begin{equation}
\mathrm{RIS}(I_t, I_{t+1}, I_{gt}) =
\begin{cases}
\dfrac{s_{t+1} - s_t}{1 - s_t}, & s_{t+1} > s_t, \\[6pt]
0, & \text{otherwise.}
\end{cases}
\label{eq:ris}
\end{equation}
Compared with absolute similarity metrics, RIS captures relative progress across consecutive renderings, providing a stable and scale-invariant indicator of refinement effectiveness.

\begin{table}[t]
\centering
\small
\setlength{\tabcolsep}{4pt}
\renewcommand{\arraystretch}{1.1}
\caption{
Structural statistics of the \textbf{VisDiffUI} dataset. 
A \cmark\ indicates paired supervision between visual differences and code edits, 
while \xmark\ denotes unpaired samples used for self-refinement and evaluation. 
\textit{Token} refers to average code length, \textit{DOM} to document depth, and \textit{Block} to layout component count.
}
\label{tab:dataset-avg-structural}
\begin{tabular}{lcccccc}
\toprule
\textbf{Partition} & \textbf{Paired} & \textbf{Samples} & \textbf{Token} & \textbf{DOM} & \textbf{Block} \\
\midrule
SFT   & \cmark & 20K  & 1.95K & 8.3  & 16.4 \\
RL(GRPO)    & \xmark & 400  & 2.77K & 10.2 & 28.1 \\
EVAL  & \xmark & 100  & 2.08K & 8.4  & 16.0 \\
\bottomrule
\end{tabular}
\end{table}

\paragraph{Reward Design.}
To transform visual differences into actionable learning feedback, we design a composite reward consisting of three components:
\begin{itemize}
    \item \textbf{Format reward} $r_{\text{format}}$: penalizes syntactically invalid HTML or CSS with $-1$, and gives $0$ otherwise;
    \item \textbf{Refinement reward} $r_{\text{improve}}$: assigns $1$ if the new rendering improves visual similarity ($s_{t+1} > s_t$), and $0$ otherwise;
    \item \textbf{Quality reward} $r_{\text{quality}}$: uses the continuous RIS value to measure the degree of improvement.
\end{itemize}
The total reward is given by:
\begin{equation}
r_t = r_{\text{format}} + r_{\text{improve}} + r_{\text{quality}}.
\label{eq:reward}
\end{equation}
This hierarchical reward structure first ensures syntactic validity, then emphasizes consistent visual enhancement, and finally encourages sensitivity to subtle perceptual differences.

\paragraph{Optimization with GRPO.}
We employ Group Relative Policy Optimization (GRPO) to enable stable learning under visual feedback.  
For each target screenshot $I_{gt}$, the model generates a candidate implementation $C_t$, renders it as $I_t$, and samples $G$ refined code variants $\{C^{(i)}\}_{i=1}^{G}$ from the current policy.  
Each variant is rendered to obtain $\{I^{(i)}\}$, and its reward $r_i$ is computed according to Eq.~\eqref{eq:reward}.  
Within each group, the normalized advantage is calculated as:
\begin{equation}
\hat{r}_i = \frac{r_i - \bar{r}}{\sigma_r + \varepsilon},
\label{eq:adv}
\end{equation}
where $\bar{r}$ and $\sigma_r$ represent the mean and deviation of rewards within the group.  
The policy is updated by minimizing the GRPO objective:
\begin{equation}
\mathcal{L}_{\text{GRPO}} = - \frac{1}{G} \sum_{i=1}^{G}
\min\!\bigl(\rho_i \hat{r}_i,\;
\mathrm{clip}(\rho_i, 1 - \epsilon, 1 + \epsilon)\hat{r}_i\bigr),
\label{eq:grpo}
\end{equation}
where $\rho_i = 
\dfrac{\pi_{\theta}(C^{(i)} \mid I_{gt}, I_t, C_t)}{\pi_{\theta_{\text{old}}}(C^{(i)} \mid I_{gt}, I_t, C_t)}$.  
Each epoch begins by generating new single-step refinements from the current checkpoint, ensuring that the model continually learns from its own most recent predictions.  
Through this quantitative difference optimization, VisRefiner progressively strengthens its ability to minimize visual discrepancies, achieving improved visual fidelity and more consistent structural alignment in generated UIs.

\begin{table*}[t]
\centering
\small
\setlength{\tabcolsep}{3pt}
\renewcommand{\arraystretch}{1.10}

\caption{
Comparison of \textbf{single-step} and \textbf{self-refinement} across
\textbf{Design2Code}, \textbf{Design2Code-HARD}, and \textbf{VisDiffUI-TEST}.
All metrics and AVG are reported with one decimal place. Each block highlights
the best (\textbf{bold}) and second-best (\underline{underline}) model within
each dataset/mode. $\Delta$ indicates the AVG improvement under self-refinement.
}
\label{tab:main_all}

\begin{tabularx}{\textwidth}{
    p{2.6cm}
    *{5}{>{\centering\arraybackslash}X}
    >{\centering\arraybackslash}X |
    *{5}{>{\centering\arraybackslash}X}
    >{\centering\arraybackslash}X |
    >{\centering\arraybackslash}p{0.9cm}
}
\toprule

\multirow{2}{*}{\textbf{Model}} &
\multicolumn{6}{c||}{\cellcolor{red!12}\textbf{Single-step}} &
\multicolumn{6}{c|}{\cellcolor{blue!12}\textbf{Self-refinement}} &
\multirow{2}{*}{\textbf{$\Delta$}} \\
\cmidrule(lr){2-7} \cmidrule(lr){8-13}
& Block & Text & Pos. & Color & CLIP & \textbf{AVG}
& Block & Text & Pos. & Color & CLIP & \textbf{AVG}
& \\

\midrule\midrule
\multicolumn{14}{c}{\textbf{Design2Code}} \\
\midrule

GPT-4o
& \textbf{93.0} & \textbf{98.2} & \textbf{85.5} & \textbf{84.1} & \textbf{90.4} & \textbf{90.2}
& \textbf{92.7} & \textbf{98.6} & \textbf{84.9} & \underline{83.3} & \underline{90.1} & \textbf{89.9}
& $-$0.3 \\

GPT-4V
& 85.8 & 97.4 & 80.5 & 73.3 & 86.9 & 84.8
& 88.8 & 98.1 & 81.1 & 72.9 & 87.2 & 85.6
& +0.8 \\

Claude 3 Opus
& 90.2 & 97.5 & 77.9 & 71.4 & 87.0 & 84.8
& 90.3 & 98.1 & 78.1 & 69.7 & 86.6 & 84.6
& $-$0.2 \\

Qwen2.5-VL-7B
& 88.2 & \underline{98.0} & 79.8 & 73.5 & 87.7 & 85.4
& 86.8 & 98.0 & 79.0 & 72.9 & 87.7 & 84.9
& $-$0.5 \\

MiMo-VL-7B-RL
& 87.8 & 97.0 & 81.1 & 77.5 & 84.7 & 85.6
& 89.0 & 97.2 & 79.5 & 77.4 & 84.6 & 85.5
& $-$0.1 \\

Qwen2.5-VL-72B
& \underline{90.7} & 97.8 & \underline{84.0} & 81.9 & 86.6 & 88.2
& 91.1 & 97.7 & \underline{84.7} & 82.3 & 87.0 & 88.6
& +0.4 \\

\rowcolor{gray!10}
\textbf{VisRefiner-7B}
& 90.6 & 97.5 & 81.7 & \underline{83.1} & \underline{90.1} & \underline{88.6}
& \underline{91.4} & \underline{98.1} & 82.3 & \textbf{83.8} & \textbf{90.7} & \underline{89.3}
& +0.7 \\

\midrule\midrule
\multicolumn{14}{c}{\textbf{Design2Code-HARD}} \\
\midrule

GPT-4o
& 56.6 & 89.8 & 78.6 & 81.9 & 87.1 & 78.8
& 72.1 & \underline{96.4} & 81.1 & 82.4 & \underline{88.2} & 84.0
& +5.2 \\

Claude 3.5 Sonnet
& 61.7 & 91.1 & \textbf{83.0} & \textbf{84.4} & \textbf{89.5} & \underline{81.9}
& 71.9 & \textbf{96.5} & \textbf{82.6} & \underline{83.0} & \textbf{88.8} & \underline{84.6}
& +2.7 \\

Claude 3 Opus
& 57.1 & 88.7 & 74.2 & 72.4 & 85.8 & 75.6
& \underline{73.3} & 95.9 & 76.6 & 70.0 & 85.6 & 80.3
& +4.7 \\

Qwen2.5-VL-7B
& 62.2 & 91.4 & 73.0 & 74.3 & 82.8 & 76.7
& 63.2 & 92.6 & 74.0 & 76.1 & 83.2 & 77.8
& +1.1 \\

MiMo-VL-7B-RL
& 61.1 & 91.0 & 77.8 & 77.5 & 84.5 & 78.4
& 62.3 & 91.9 & 77.1 & 77.9 & 84.4 & 78.7
& +0.3 \\

Qwen2.5-VL-72B
& \underline{64.9} & \underline{94.8} & 79.2 & 80.4 & 84.8 & 80.8
& 72.8 & 94.9 & 79.8 & 80.9 & 84.9 & 82.7
& +1.9 \\

\rowcolor{gray!10}
\textbf{VisRefiner-7B}
& \textbf{78.1} & \textbf{96.2} & \underline{82.3} & \underline{83.7} & \underline{87.9} & \textbf{85.6}
& \textbf{80.8} & 96.2 & \underline{82.2} & \textbf{84.8} & 87.7 & \textbf{86.3}
& +0.7 \\

\midrule\midrule
\multicolumn{14}{c}{\textbf{VisDiffUI-TEST}} \\
\midrule

Qwen2.5-VL-7B
& 91.9 & 98.8 & 80.3 & 79.2 & 87.2 & 87.5
& 91.5 & 97.9 & 79.6 & 78.1 & 87.0 & 86.8
& $-$0.7 \\

MiMo-VL-7B-RL
& 85.6 & 94.9 & 81.6 & 77.7 & 86.3 & 85.2
& 87.4 & 94.5 & 79.7 & 76.3 & 85.9 & 84.8
& $-$0.4 \\

Qwen2.5-VL-72B
& \textbf{95.2} & \underline{99.0} & \textbf{87.4} & \underline{85.4} & \underline{88.1} & \underline{91.0}
& \underline{95.4} & \underline{99.0} & \textbf{87.3} & \underline{86.1} & \underline{88.6} & \underline{91.3}
& +0.3 \\

\rowcolor{gray!10}
\textbf{VisRefiner-7B}
& \underline{93.9} & \textbf{99.3} & \underline{86.9} & \textbf{89.3} & \textbf{90.5} & \textbf{92.0}
& \textbf{96.6} & \textbf{99.1} & \textbf{87.3} & \textbf{89.7} & \textbf{90.0} & \textbf{92.5}
& +0.5 \\

\bottomrule
\end{tabularx}

\end{table*}

\section{Experiments}
\label{sec:exp}

\subsection{Implementation Details}

\paragraph{Datasets.}
We construct \textbf{VisDiffUI}, a difference-aligned dataset for screenshot-to-code generation, derived from large-scale public UI-to-code sources Vision2UI~\cite{gui2024vision2ui} and WebSight~\cite{laurenccon2024unlocking}. 
All raw HTML/CSS samples are standardized through a unified preprocessing pipeline that removes invalid markup, replaces external assets with size-preserving placeholders, and inlines all styles to ensure consistent rendering. 
We further apply quality filtering using Qwen2.5-Coder-32B-Instruct~\cite{hui2024qwen2} to verify syntactic correctness and visual fidelity, resulting in 7K fully renderable code–image pairs. 
The construction of difference-aligned supervision follows the procedure described in Section~\ref{sec:method}, generating approximately 20K training instances with aligned visual and code changes. 
The dataset is divided into three partitions: paired samples for supervised difference-aligned training, a smaller unpaired subset with higher structural complexity for reinforcement learning based self-refinement, and an evaluation set with average complexity for benchmarking. 
Table~\ref{tab:dataset-avg-structural} summarizes the structural statistics of each partition.

\paragraph{Training.}
We adopt \textbf{Qwen2.5-VL-7B-Instruct}~\cite{bai2025qwen2} as the backbone model. 
Supervised fine-tuning is first conducted on the paired subset of \textbf{VisDiffUI} for one epoch with a batch size of 64 and a learning rate of $1\times10^{-5}$. 
Reinforcement learning based on GRPO\cite{shao2024deepseekmath} is then performed on the unpaired subset containing 400 structurally complex samples for eight epochs, with a rollout size of 8 and a learning rate of $2\times10^{-6}$. 

\paragraph{Evaluation.}
Evaluation is carried out on three datasets:  
(1) the \textbf{Design2Code} benchmark~\cite{si2024design2code}, which includes 484 real-world UI screenshots paired with reference HTML;  
(2) the \textbf{Design2Code-HARD} variant containing denser and more complex layouts; and  
(3) our held-out \textbf{VisDiffUI-Test} set with 100 diverse examples of average complexity.  

Following the official Design2Code protocol, we adopt five fine-grained metrics that jointly evaluate layout fidelity, textual accuracy, spatial alignment, stylistic consistency, and perceptual similarity:
\begin{itemize}
  \item \textbf{Block Match}: the proportion of correctly aligned visual block regions between generated and reference renderings.
  \item \textbf{Text Accuracy}: the character-level Sørensen-Dice coefficient over matched text regions.
  \item \textbf{Position Alignment}: the normalized spatial offset between corresponding block centers.
  \item \textbf{Color Consistency}: the mean perceptual similarity measured by the CIEDE2000 color difference.
  \item \textbf{CLIP Similarity}: the perceptual alignment of masked renderings computed using CLIP-ViT-B/32 embeddings, where text regions are inpainted to ensure layout-based comparison.
\end{itemize}
Each metric captures a distinct dimension of visual fidelity. Following prior work, we report all submetrics independently rather than aggregating them into a single composite score.

\paragraph{Baselines.}
We compare \textbf{VisRefiner-7B} with both proprietary and open-source multimodal large language models under \textit{single-step} and \textit{self-refinement} inference modes. 
Proprietary models include GPT-4o, GPT-4V, Claude 3.5 Sonnet, and Claude 3 Opus, with results taken from the Design2Code benchmark~\cite{si2024design2code}. 
Open-source comparisons include Qwen2.5-VL-7B, Qwen2.5-VL-72B, and MiMo-VL-7B-RL~\cite{coreteam2025mimovltechnicalreport}. 
All models are evaluated under identical rendering and scoring protocols to ensure fairness.

\subsection{Main Results}

Table~\ref{tab:main_all} summarizes results on the Design2Code and Design2Code-HARD benchmarks, together with our held-out VisDiffUI-Test set. All models are evaluated under both single-step generation and one-step self-refinement. We highlight several key observations.

\paragraph{Strong single-step performance.}
VisRefiner-7B exhibits consistently strong single-step accuracy across all benchmarks. 
On Design2Code, it reaches an average score of 88.6, close to GPT-4o (90.2) while surpassing open-source models including Qwen2.5-VL-72B (88.2). 
On the more challenging Design2Code-HARD split, VisRefiner achieves the highest overall single-step performance with an average of 85.6. 
It provides notable advantages in structural and textual alignment, improving Block Match by 21.5 points over GPT-4o (78.1 vs.\ 56.6) and Text Accuracy by 6.4 (96.2 vs.\ 89.8). 
These results show that visually grounded, refinement-oriented training substantially enhances layout reasoning and text fidelity, even at the 7B scale.

\paragraph{Refinement improves stability rather than fluctuating.}
Across baselines, refinement often produces inconsistent behavior. 
GPT-4o increases by +5.2 points on the HARD split (84.0 vs.\ 78.8) but decreases on the standard benchmark (89.9 vs.\ 90.2). 
Open-source models such as Qwen2.5-VL-7B (+1.1 / --0.5) and MiMo-VL-7B-RL (+0.3 / --0.1) show similarly irregular trends, indicating that refinement is not reliably activated in general-purpose vision-language models. 
VisRefiner, in contrast, shows consistent gains across all datasets, including held-out in-domain samples from VisDiffUI-Test, improving by +0.7 on Design2Code, +0.7 on Design2Code-HARD, and +0.5 on VisDiffUI-Test. 
The uniform improvements across public and held-out data suggest that VisRefiner internalizes refinement behavior during training and applies corrections in a stable, generalizable manner rather than relying on stochastic or overly conservative adjustments.


\section{Analysis}
\label{sec:analysis}

\begin{table}[t]
\centering
\small
\setlength{\tabcolsep}{2.8pt}
\renewcommand{\arraystretch}{1.12}

\caption{
Comparison of training strategies on VisDiffUI-Test under the single-step setting.
}
\label{tab:strategy_results}

\begin{tabular}{lccccc|c}
\toprule
\textbf{Model Variant} & Block & Text & Pos. & Color & CLIP & \textbf{AVG} \\
\midrule
Base (Qwen2.5-VL-7B) & 91.9 & 98.8 & 80.3 & 79.2 & 87.2 & 87.5 \\
\rowcolor{gray!8}
\quad + GRPO & 92.6 & 98.9 & 80.2 & 81.1 & 87.3 & 88.0 \\
\rowcolor{gray!8}
\quad + SFT & 93.2 & 98.3 & 84.2 & 88.4 & 89.9 & 90.8 \\
\rowcolor{gray!12}
\quad + SFT + GRPO & \textbf{93.9} & \textbf{99.3} & \textbf{86.9} & \textbf{89.3} & \textbf{90.5} & \textbf{92.0} \\
\bottomrule
\end{tabular}
\end{table}

\subsection{Effect of Training Strategy}

\textbf{Training variants and overall trend.}
Table~\ref{tab:strategy_results} ablates our training pipeline on VisDiffUI-Test under the single-step setting. 
We begin with the base Qwen2.5-VL-7B model, which achieves an average score of 87.5. 
Introducing GRPO, then SFT, and finally the combination of SFT and GRPO results in steady improvements up to 92.0.  
Although all variants are trained with refinement-style data, their single-step accuracy increases consistently, showing that refinement-oriented supervision enhances forward generation even without explicit refinement during inference.

\textbf{SFT as the main driver of structural and visual quality.}
Applying GRPO alone leads to only a modest increase from 87.5 to 88.0 and largely preserves the behavior of the base model.  
In contrast, supervised fine-tuning produces the most substantial improvement, raising the average to 90.8.  
SFT significantly improves positional alignment (from 80.3 to 84.2) and color consistency (from 79.2 to 88.4), and also strengthens CLIP similarity.  
These findings indicate that difference-aligned SFT is the primary factor that strengthens the model’s ability to capture spatial and visual discrepancies.

\textbf{GRPO benefits from the difference-aware foundation built by SFT.}
Adding GRPO after SFT yields the best overall performance, reaching 92.0 on average.  
Although GRPO is applied using only 400 examples, it still improves every fine-grained metric, including Block (93.2 to 93.9), Text (98.3 to 99.3), and Position (84.2 to 86.9).  
This pattern suggests that GRPO is most effective when the model has already learned a rich difference-aware representation from SFT, allowing GRPO to further consolidate fine-grained judgment and produce a stronger refinement-oriented model.

\begin{figure}[t]
    \centering

    \begin{subfigure}{0.95\linewidth}
        \centering
        \includegraphics[width=\linewidth]{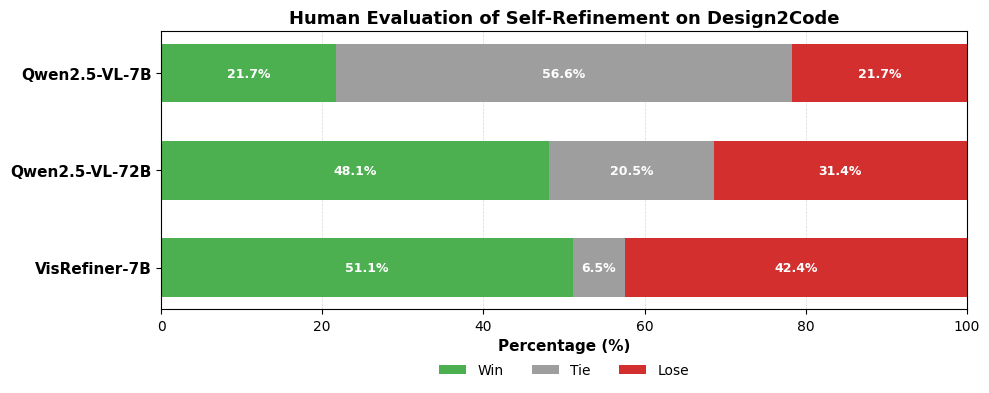}
    \end{subfigure}
    \vskip 6pt

    \begin{subfigure}{0.95\linewidth}
        \centering
        \includegraphics[width=\linewidth]{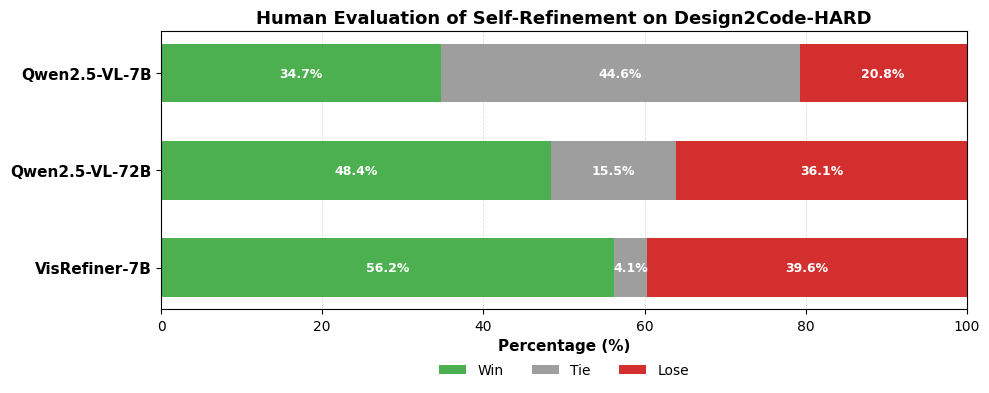}
    \end{subfigure}

    \caption{
   \textbf{ Human evaluation} of one-step self-refinement. 
    Annotators compared each model’s refined output against its own single-step generation, labeling each comparison as Win (improved), Tie (unchanged), or Lose (degraded).
    }
    \label{fig:human-eval}
\end{figure}

\subsection{Human Evaluation of Refinement Behavior}

\paragraph{Setup.}
We conduct a human evaluation on both Design2Code and Design2Code-HARD to assess whether refinement meaningfully improves generation quality. 
Annotators compare each model’s refined output with its own single-step prediction and assign a Win, Tie, or Lose label, indicating improvement, no visible change, or degradation.

\textbf{Visual refinement as a higher-order capability.}
Refining one’s own output requires the model not only to generate code, but also to identify potential errors, interpret visual discrepancies, and propose targeted edits. 
This makes refinement a higher-order capability that does not naturally emerge in smaller MLLMs. 
As shown in Figure~\ref{fig:human-eval}, the base Qwen2.5-VL-7B model exhibits very limited refinement behavior: nearly half of all samples are labeled as Tie, meaning the model often produces outputs that are visually indistinguishable from its single-step predictions. 
This suggests that the model lacks mechanisms to detect its own mistakes or decide where a revision is needed.

\textbf{Larger models partially acquire this ability.}
Increasing model capacity helps alleviate this limitation. 
Qwen2.5-VL-72B shows fewer Ties and more Wins, indicating that larger MLLMs develop a stronger ability to recognize errors and apply small corrective edits. 
However, a substantial number of refinement attempts still fail to produce noticeable improvements, highlighting that capacity alone provides only a partial solution and does not reliably induce robust refinement behavior.

\textbf{VisRefiner learns deliberate and active refinement.}
VisRefiner-7B demonstrates a markedly different profile. 
Its Tie rate is significantly lower than both Qwen models, showing that it consistently attempts nontrivial revisions rather than defaulting to no-op predictions. 
This behavior aligns with its training objective, which explicitly exposes the model to iterative correction signals. 
The model’s more active refinement strategy results in a slightly higher Lose rate, reflecting that larger edits occasionally degrade output quality, but also confirming that the model is actively engaging in correction rather than behaving conservatively.

These findings indicate that refinement is a structured capability that does not reliably emerge in standard MLLMs, especially at smaller scales.

\begin{table}[t]
\centering
\small
\setlength{\tabcolsep}{2.3pt}
\renewcommand{\arraystretch}{1.12}

\caption{
\textbf{Multi-turn self-refinement results on \textsc{VisDiffUI-Test}.}
For each turn, we report the current average score (\textit{Current}) and the best result across all previous turns (\textit{Filtered}).
}
\label{tab:refine_turns}

\begin{tabular}{l|cc|cc|cc|cc}
\toprule
\multirow{2}{*}{\textbf{Model}} &
\multicolumn{2}{c|}{\textbf{Turn 0}} &
\multicolumn{2}{c|}{\textbf{Turn 1}} &
\multicolumn{2}{c|}{\textbf{Turn 3}} &
\multicolumn{2}{c}{\textbf{Turn 5}} \\
\cmidrule(lr){2-9}
& \textit{Curr.} & \textit{Filt.} & \textit{Curr.} & \textit{Filt.} &
  \textit{Curr.} & \textit{Filt.} & \textit{Curr.} & \textit{Filt.} \\
\midrule

\rowcolor{gray!4}
Qwen2.5-VL-7B &
87.5 & 87.5 &
86.8 & 87.7 &
86.6 & 87.9 &
86.6 & 87.9 \\

\rowcolor{gray!10}
\textbf{VisRefiner-7B} &
\textbf{92.0} & \textbf{92.0} &
\textbf{92.5} & \textbf{93.2} &
\textbf{92.9} & \textbf{93.5} &
\textbf{92.6} & \textbf{94.2} \\

\bottomrule
\end{tabular}
\end{table}

\subsection{Multi-turn Self-Refinement Analysis}

Table~\ref{tab:refine_turns} analyzes model behavior under multi-turn self-refinement on \textsc{VisDiffUI-Test}.  
At each turn, the model takes its previous output as input and regenerates the code.  
We report both the performance of the current turn (\textit{Current}) and the best result across all turns (\textit{Filtered}), where the latter simulates a simple selection strategy based on visual similarity metrics such as CLIP score.

\paragraph{Effect of iterative refinement.}  
VisRefiner-7B achieves strong single-turn performance, while the results begin to fluctuate as the number of refinement turns increases.  
This indicates that additional iterations do not consistently yield better quality once major discrepancies have been corrected.  
However, the filtered results continue to improve, reaching 94.2 at Turn~5, suggesting that intermediate outputs still contain valuable alternatives.  
Such diversity implies that VisRefiner explores multiple correction paths during refinement, even though it is primarily optimized for single-step improvement.  
Compared with Qwen2.5-VL-7B, which exhibits unstable or degrading performance across turns, VisRefiner maintains stability and shows greater potential when combined with simple filtering or selection strategies.  
These findings suggest that difference-driven learning not only enhances the model’s one-step precision but also enriches its exploration space across turns.  

\section{Conclusion}
\label{sec:conclusion}

We present \textbf{VisRefiner}, a difference-driven training framework that enables multimodal large language models to improve screenshot-to-code generation by learning directly from visual discrepancies. 
Instead of treating visual differences as post-hoc feedback, VisRefiner incorporates them into the training process, allowing the model to understand how code modifications influence visual outcomes and to refine its predictions accordingly. 
The framework combines difference-aligned supervision with GRPO-based self-refinement, helping the model connect visual appearance with implementation details and produce more faithful and visually consistent UI code. 
Extensive experiments demonstrate that VisRefiner enhances visual fidelity, structural alignment, and self-improvement capability. 
By transforming visual differences into a direct learning signal, VisRefiner moves multimodal reasoning toward a more human-like paradigm of perception-guided code generation.

\clearpage

{
    \small
    \bibliographystyle{ieeenat_fullname}
    \bibliography{main}
}

\clearpage
\setcounter{page}{1}

\section*{GRPO Training Procedure}
\label{app:grpo}

This section provides the full training procedure for the reinforcement stage of \textbf{VisRefiner}. 
We use Group Relative Policy Optimization (GRPO) to allow the model to refine its own generated code based on quantitative visual differences. 
Each iteration consists of four steps: generating candidate refinements, rendering visual outputs, computing perceptual rewards, and updating the policy using group-wise relative advantages.

\paragraph{Reward Function.}
For a rendered prediction $I_t$ and its refined version $I_{t+1}$, we compute a CLIP-based similarity score $s(\cdot)$ between each image and the target screenshot $I_{gt}$.  
The reward combines structural validity, improvement detection, and the magnitude of visual refinement:

\[
r = r_{\text{format}} + r_{\text{improve}} + r_{\text{quality}},
\]

where  
\[
r_{\text{format}} =
\begin{cases}
-1, & \text{if the generated HTML/CSS is invalid}, \\
0, & \text{otherwise},
\end{cases}
\]

\[
r_{\text{improve}} =
\begin{cases}
1, & s(I_{t+1}) > s(I_t), \\
0, & \text{otherwise},
\end{cases}
\]

\[
r_{\text{quality}} =
\begin{cases}
\dfrac{s(I_{t+1}) - s(I_t)}{1 - s(I_t)}, & s(I_{t+1}) > s(I_t), \\
0, & \text{otherwise}.
\end{cases}
\]

This formulation rewards both correctness and the degree of visual improvement, while maintaining stability across varying similarity scales.

\paragraph{Group-wise Advantage.}
Given a group of rewards $\{r_i\}_{i=1}^G$, we compute the normalized advantage:

\[
\hat{r}_i = \frac{r_i - \bar{r}}{\sigma_r + \varepsilon},
\]

where $\bar{r}$ and $\sigma_r$ are the mean and standard deviation within the group.

\paragraph{GRPO Objective.}
For each sampled refinement $C^{(i)}$, the policy ratio is

\[
\rho_i = \frac{
\pi_\theta(C^{(i)} \mid I_{gt}, I_t, C_t)
}{
\pi_{\theta_{\text{old}}}(C^{(i)} \mid I_{gt}, I_t, C_t)
}.
\]

The GRPO loss is

\[
\mathcal{L}_{\text{GRPO}} = - \frac{1}{G} \sum_{i=1}^G 
\min\!\left(
\rho_i \hat{r}_i,\;
\text{clip}(\rho_i, 1 - \epsilon, 1 + \epsilon)\hat{r}_i
\right).
\]

Here the term $\text{clip}(\cdot)$ is the standard PPO-style clipping
operation and is unrelated to the CLIP model used for similarity scoring. More implementation details are provided in Section \ref{sec:appendix_exp}.

\begin{algorithm}[t]
\caption{VisRefiner: GRPO with Self-Refinement}
\label{alg:grpo_appendix}
\begin{algorithmic}[1]
\State \textbf{Input:} target screenshots $\mathcal{I}_{gt}$, SFT-initialized policy $\pi_{\theta_0}$, group size $G$, epochs $E$
\For{$e = 0$ to $E-1$}
    \State $\pi_{\theta_{\text{old}}} \gets \pi_{\theta_e}$
    \State $\mathcal{D} \gets \emptyset$
    \ForAll{$I_{gt} \in \mathcal{I}_{gt}$}
        \State $C_t \sim \pi_{\theta_{\text{old}}}(I_{gt})$
        \State $I_t \gets \mathrm{Render}(C_t)$
        \State $\mathcal{D} \gets \mathcal{D} \cup \{(I_{gt}, I_t, C_t)\}$
    \EndFor
    \ForAll{$(I_{gt}, I_t, C_t) \in \mathcal{D}$}
        \State Sample $G$ refinements $\{C^{(i)}\}$
        \For{$i = 1$ to $G$}
            \State $I^{(i)} \gets \mathrm{Render}(C^{(i)})$
            \State Compute reward $r_i$
        \EndFor
        \State Compute advantages $\{\hat{r}_i\}$
        \State Compute policy ratios $\{\rho_i\}$
        \State Update $\theta_e$ with GRPO loss
    \EndFor
\EndFor
\State \textbf{Output:} final policy $\pi_{\theta_E}$
\end{algorithmic}
\end{algorithm}

\begin{figure*}[!t]
    \centering
    \includegraphics[width=1.0\linewidth]{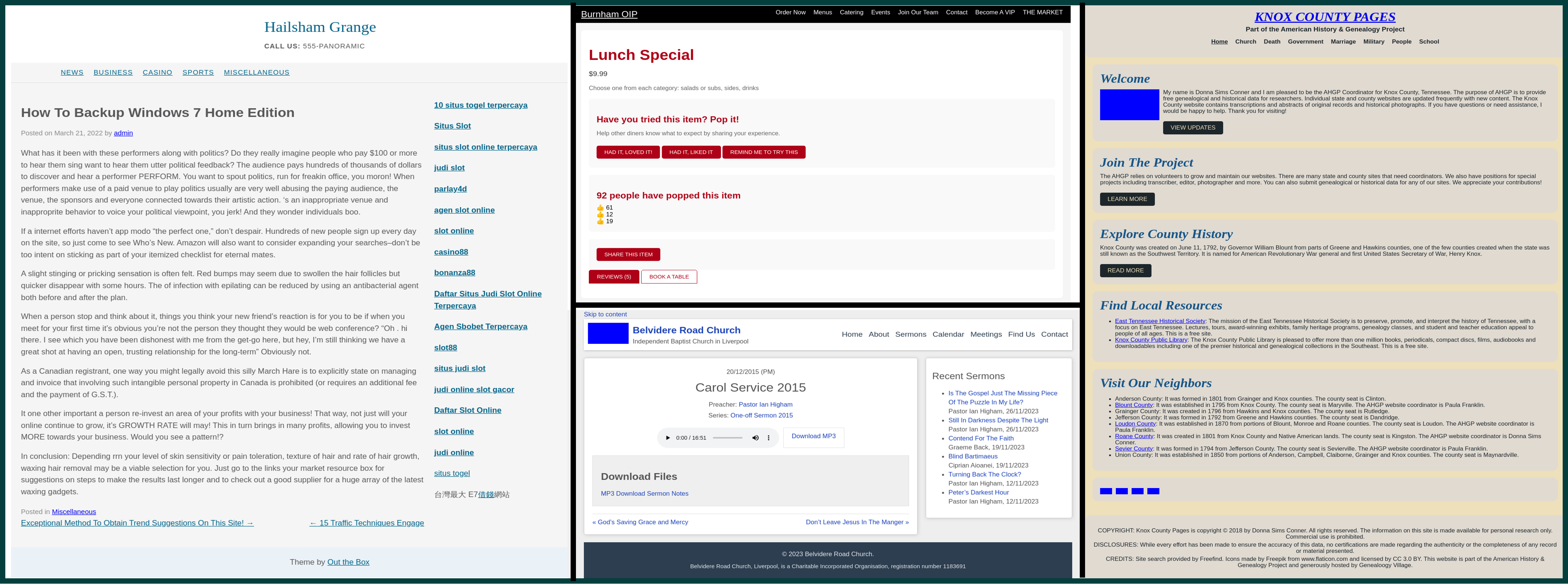}
    \caption{Representative examples from the seed version of \textsc{VisDiffUI}, covering diverse layouts and component structures.}
    \label{fig:example_VisDiffUI}
\end{figure*}

\section{VisDiffUI Dataset}
\label{sec:dataset}

\subsection{Data Preprocessing and Normalization}

The construction of \textbf{VisDiffUI Dataset} begins with a comprehensive preprocessing pipeline designed to produce clean, self-contained, and consistently renderable HTML samples. This ensures reliable training for visual refinement and layout-aware code generation.

\textbf{Self-containment and renderability.}  
We first filter out HTML samples that depend on external resources such as remote images, fonts, or scripts, which frequently cause broken renderings or unstable layouts. Only files that can be rendered independently in a local browser environment are retained.

\textbf{Image placeholder substitution.}  
To eliminate broken or inconsistent image links, all external URLs are replaced with uniform local placeholders, while preserving the original \texttt{width} and \texttt{height} attributes.  
This maintains layout fidelity and allows the model to focus on learning alignment, spacing, and structure rather than irrelevant image content.

\textbf{HTML sanitization and validation.}  
We remove incomplete markup, broken tags, and script-heavy elements, then verify renderability via headless browser execution.  
Samples that fail to produce valid visual outputs are discarded, ensuring a high-quality, fully visual dataset.

\textbf{Generative normalization.}  
To improve structural consistency, each HTML file is reformatted through a generative rewriting process using the \textbf{Qwen2.5-Coder-32B-Instruct} model.  
This step merges internal CSS, removes redundancy, and standardizes tag usage while preserving original appearance, yielding a cleaner and more consistent layout representation.

The result is a collection of HTML–image pairs that are visually coherent, structurally standardized, and renderable in isolation.  
We refer to this as the \textit{seed version} of VisDiffUI, which serves as the foundation for downstream annotation and training (see Figure~\ref{fig:example_VisDiffUI}).

\subsection{GRPO Subset Construction}

\begin{figure}[t]
    \centering
    \includegraphics[width=1.0\linewidth]{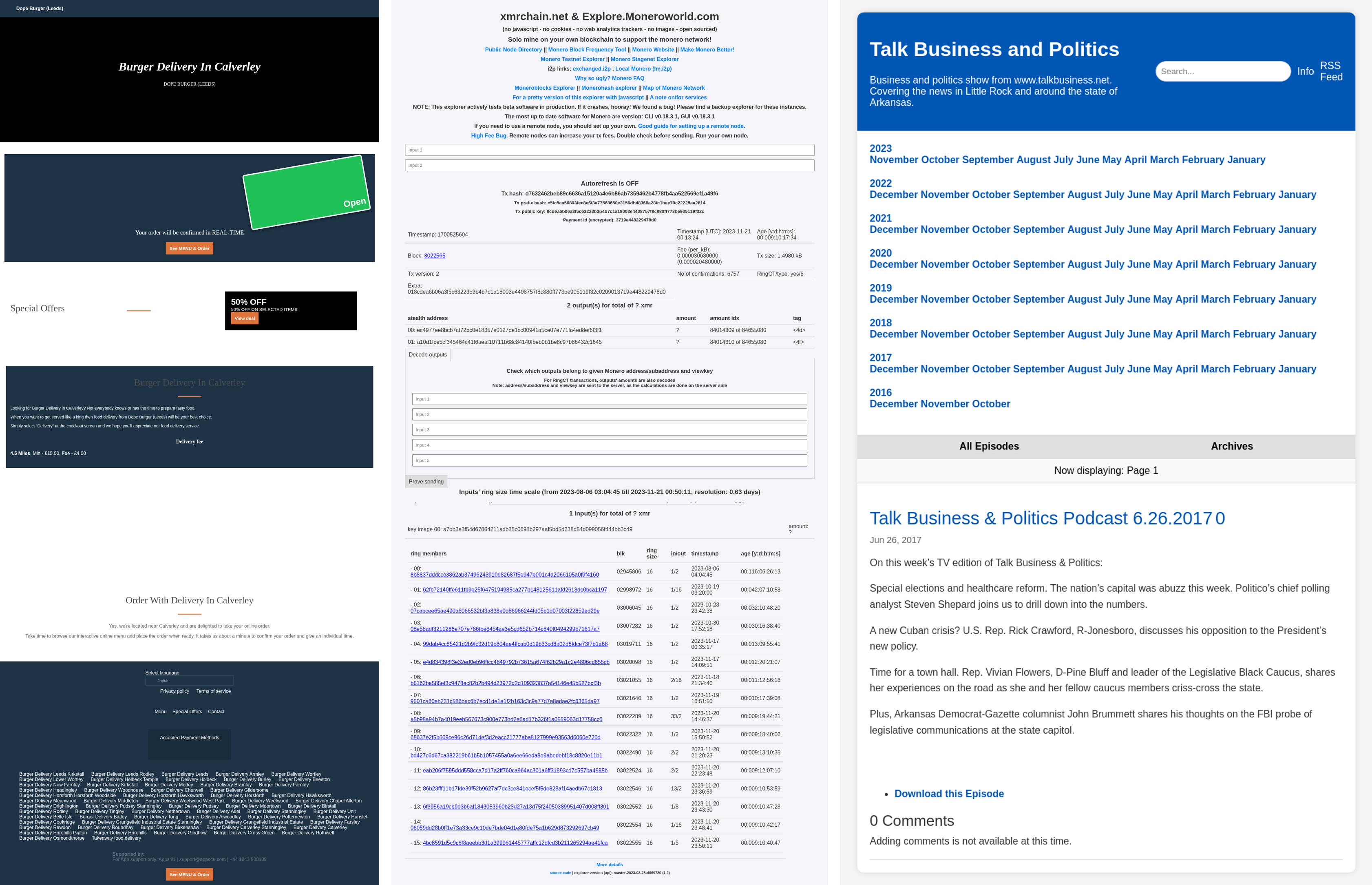}
    \caption{Samples from the GRPO subset, featuring complex visual structures and hierarchical layouts used for reinforcement learning.}
    \label{fig:grpo_examples}
\end{figure}

To provide strong visual learning signals for reinforcement training, we construct a curated subset of 400 challenging examples, referred to as the \textbf{GRPO subset} (Figure~\ref{fig:grpo_examples}).  
These samples emphasize complex structure, deep DOM hierarchies, and rich layout interactions, making them particularly suitable for refinement-based optimization.

Sample difficulty is estimated through a composite score derived from structural and stylistic features, including DOM depth, tag diversity, inline style count, and script density.  
For each sample $i$, we compute a normalized feature vector $x_i \in \mathbb{R}^d$ and define its difficulty as:
\[
s_i = \sum_{j=1}^d \frac{x_{ij} - \mu_j}{\sigma_j},
\]
where $\mu_j$ and $\sigma_j$ denote the dataset mean and standard deviation.  
We select the top 25\% of samples by difficulty and randomly draw 400 for GRPO training, excluding any overlap with the test set.  
The test set itself contains 100 examples evenly distributed across the difficulty spectrum to ensure balanced evaluation.

This design ensures that the GRPO subset provides diverse and high-gradient supervision, enabling efficient reinforcement learning even under limited data size and promoting robust generalization to unseen UI patterns.

\subsection{Rule-Based Perturbation Specifications}
\label{sec:appendix_perturb}

To construct difference-aligned refinement pairs, we apply a set of
localized and render-preserving perturbation rules to clean HTML–CSS
implementations.  
Each rule introduces a controlled discrepancy that produces a visually
observable change while keeping the perturbed code valid and renderable.
Below we provide compact before–after examples for the six perturbation
categories used in our dataset.

\vspace{0.5em}
\noindent\textbf{Color Perturbations.}  
Modify palette tokens or inline color attributes to create hue shifts.

\begin{tcolorbox}[
  colback=gray!2,
  colframe=black!20,
  boxrule=0.2pt,
  left=2pt,right=2pt,top=2pt,bottom=2pt]
\scriptsize
\textbf{Original:}
\begin{verbatim}
<button style="background:#4A90E2">OK</button>
\end{verbatim}
\textbf{Perturbed:}
\begin{verbatim}
<button style="background:#C74AE2">OK</button>
\end{verbatim}
\end{tcolorbox}

\vspace{0.3em}
\noindent\textbf{Layout Perturbations.}  
Alter flex/grid settings to create spacing or structural shifts.

\begin{tcolorbox}[
  colback=gray!2,colframe=black!20,boxrule=0.2pt,
  left=2pt,right=2pt,top=2pt,bottom=2pt]
\scriptsize
\textbf{Original:}
\begin{verbatim}
.row { display:flex; gap:20px; }
\end{verbatim}
\textbf{Perturbed:}
\begin{verbatim}
.row { display:flex; gap:3px; flex-direction:column; }
\end{verbatim}
\end{tcolorbox}

\vspace{0.3em}
\noindent\textbf{Alignment Perturbations.}  
Shift margins, padding, or alignment to distort spatial balance.

\begin{tcolorbox}[
  colback=gray!2,colframe=black!20,boxrule=0.2pt,
  left=2pt,right=2pt,top=2pt,bottom=2pt]
\scriptsize
\textbf{Original:}
\begin{verbatim}
<h1 style="text-align:center;margin-left:0">Title</h1>
\end{verbatim}
\textbf{Perturbed:}
\begin{verbatim}
<h1 style="text-align:left;margin-left:60px">Title</h1>
\end{verbatim}
\end{tcolorbox}

\vspace{0.3em}
\noindent\textbf{Component Perturbations.}  
Remove or replace entire UI modules.

\begin{tcolorbox}[
  colback=gray!2,colframe=black!20,boxrule=0.2pt,
  left=2pt,right=2pt,top=2pt,bottom=2pt]
\scriptsize
\textbf{Original:}
\begin{verbatim}
<nav class="navbar"> ... </nav>
\end{verbatim}
\textbf{Perturbed:}
\begin{verbatim}
<!-- <nav class="navbar">...</nav> -->
\end{verbatim}
\end{tcolorbox}

\vspace{0.3em}
\noindent\textbf{Image Perturbations.}  
Modify image container size while preserving renderability.

\begin{tcolorbox}[
  colback=gray!2,colframe=black!20,boxrule=0.2pt,
  left=2pt,right=2pt,top=2pt,bottom=2pt]
\scriptsize
\textbf{Original:}
\begin{verbatim}
<img src="p.png" width="300" height="200">
\end{verbatim}
\textbf{Perturbed:}
\begin{verbatim}
<img src="p.png" width="300" height="80">
\end{verbatim}
\end{tcolorbox}

\vspace{0.3em}
\noindent\textbf{Text Perturbations.}  
Modify font properties or truncate content.

\begin{tcolorbox}[
  colback=gray!2,colframe=black!20,boxrule=0.2pt,
  left=2pt,right=2pt,top=2pt,bottom=2pt]
\scriptsize
\textbf{Original:}
\begin{verbatim}
<p class="desc">Explore our latest collection</p>
\end{verbatim}
\textbf{Perturbed:}
\begin{verbatim}
<p class="desc">Explore our</p>
\end{verbatim}
\end{tcolorbox}

These controlled perturbations generate diverse visual discrepancies
while preserving syntactic validity, enabling precise difference-aligned
supervision for VisRefiner.

\section{Experiments Details}
\label{sec:appendix_exp}

This section provides additional details on model training, rendering
procedures, sampling configurations, and GRPO implementation used in
VisRefiner. Our experiments follow a two-stage pipeline consisting of
supervised difference-aligned learning and reinforcement-based
self-refinement.

\subsection{Training Setup}

We train VisRefiner in two stages:  
(1) supervised fine-tuning (SFT) on about 20k difference-aligned pairs,  
and (2) GRPO-based refinement training on a curated subset of 400
structurally challenging examples.  
Table~\ref{tab:train_summary} summarizes the training schedule.

\begin{table}[h]
\centering
\small
\renewcommand{\arraystretch}{1.15}
\begin{tabular}{lcccc}
\toprule
\textbf{Stage} & \textbf{Samples} & \textbf{Epochs} & \textbf{Hardware} & \textbf{Time} \\
\midrule
SFT  & 20k  & 1  & 8$\times$ 80G A100 & 6h \\
GRPO & 400  & 8  & 8$\times$ 80G A100 & 20h \\
\bottomrule
\end{tabular}
\caption{Summary of the training configuration for VisRefiner.}
\label{tab:train_summary}
\end{table}

\paragraph{Rendering and preprocessing.}
All HTML is rendered using a headless Chromium engine.  
To ensure consistent CLIP evaluation and avoid size-dependent bias,
rendered images are resized to a fixed pixel budget
(\texttt{MAX\_PIXELS = 1003520}).  
Both training stages operate in \texttt{bfloat16} precision with DeepSpeed
ZeRO-3, using a per-GPU batch size of 1 and gradient accumulation of 8.

\paragraph{Supervised fine-tuning.}
SFT uses a learning rate of $1\times10^{-5}$ with 5\% warmup.  
We train for a single epoch, since difference-aligned pairs provide dense,
localized supervision and additional passes lead to saturation.

\paragraph{GRPO refinement training.}
GRPO is trained for 8 epochs with a smaller learning rate of
$1\times10^{-6}$.  
Training interleaves generation and optimization: each GPU runs both
forward/backward passes and HTML sampling using vLLM.  
We allow two multimodal images per prompt (predicted render and target
design) and set the maximum generation length to 6000 tokens to
accommodate complex HTML layouts.

Sampling uses moderately exploratory settings:

\begin{itemize}
    \item nucleus sampling $p = 0.95$
    \item top-$k = 50$
    \item temperature $= 0.9$
    \item repetition penalty $= 1.05$
\end{itemize}

For each training instance, we sample a group of $G=8$ refinement
candidates in parallel.

Each refinement candidate is rendered and scored using a
visual reward that measures improvement relative to the previous render.
We compute CLIP similarity using the standard CLIP encoder (not the
masked-layout variant used in evaluation), since our goal is to measure
global perceptual improvement rather than purely structural similarity.

To reduce reward noise, we require a minimum similarity improvement of
$\Delta s > 0.001$ before counting a refinement as successful.  This
prevents half-pixel fluctuations or renderer-level noise from receiving
positive reward.

\section{Inference Pipeline}
\label{sec:inference}

The inference pipeline contains two stages: (1) initial HTML generation
from the target screenshot, and (2) optional multi-turn refinement where
the model iteratively improves its own outputs. This procedure mirrors
the evaluation setting used in our experiments.

\subsection{Initial Generation}

The model first receives the target UI screenshot and is instructed to
produce a complete HTML file that reproduces the visual layout. To
ensure consistent rendering across cases, all image URLs in the output
are replaced with fixed-size placeholder tokens. Only syntactically
complete and self-contained HTML files are kept for downstream
processing.

\begin{tcolorbox}[title=Initial Prompt, colback=white]
\textbf{System:} 

You are an expert web developer.

\textbf{User:}  

Please write HTML code that reproduces this design.  
Use inline style or internal CSS.  
All images must use ``[Placeholder]'' with fixed dimensions.
\end{tcolorbox}

\subsection{Multi-Turn Refinement}

After the initial prediction, the model may optionally enter a
multi-turn refinement loop.  
At each turn, it receives three inputs:

\begin{itemize}[itemsep=2pt]
    \item the current predicted HTML code (\texttt{pred\_code}),
    \item the rendered output of that code (\texttt{pred\_image}),
    \item the target reference screenshot (\texttt{GT\_image}).
\end{itemize}

The model is tasked with analyzing the visual differences and producing
an improved HTML version that reduces the gap.

\begin{tcolorbox}[title=Refinement Prompt, colback=white]
You are refining HTML/CSS code to match a target design.

\textbf{Inputs:}  

-- \textbf{GT\_image}: target screenshot  

-- \textbf{pred\_image}: rendering of current code  

-- \textbf{pred\_code}: latest HTML output

\textbf{Task:}

Compare the two images, identify mismatches, and return revised HTML
that improves layout, alignment, spacing, fonts, or color.

\textbf{Guidelines:}  

-- Rewrite large sections if necessary  

-- Preserve correct regions when possible  

-- Use inline or internal CSS only  

-- Replace all images with ``[Placeholder]'' and explicit size  

-- Output a complete and valid HTML file

\end{tcolorbox}

\subsection{Execution Logic}

Each refinement turn is executed sequentially. The model generates a new
HTML prediction given the triplet
(\texttt{GT\_image}, \texttt{pred\_image}, \texttt{pred\_code}).  
A retry mechanism is applied when the output is incomplete, empty, or
fails rendering.  
All intermediate predictions and rendered screenshots are stored for
analysis and reproducibility.

\subsection{Invalid Output Filtering}

We apply two validation checks to ensure reliable refinement:

\begin{itemize}[itemsep=2pt]
    \item \textbf{Code validity.}  
    The output must contain both opening and closing \texttt{</html>} tags.

    \item \textbf{Visual validity.}  
    The rendered page must successfully load and produce a non-blank
    image.
\end{itemize}

Only predictions passing both checks are accepted as valid updates for
subsequent refinement or evaluation.

\section{Case Study}

\begin{figure*}[p]
    \centering
    \includegraphics[width=\textwidth]{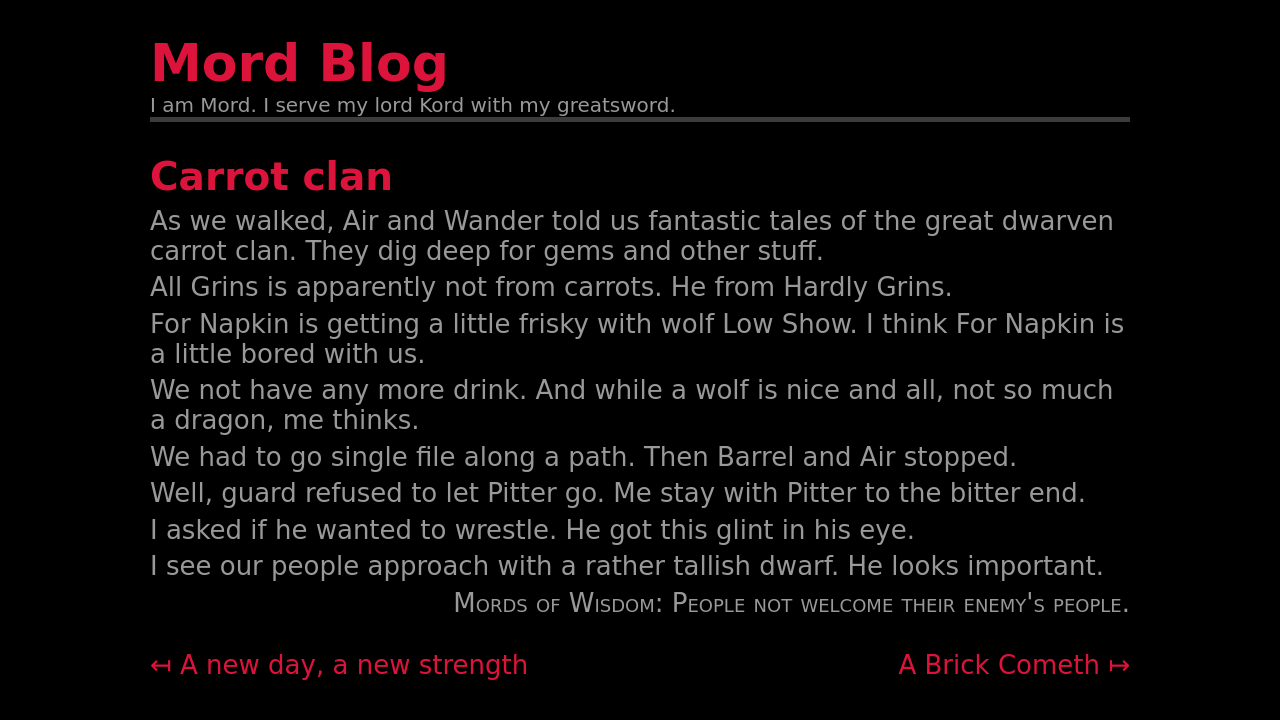}
    \caption{
        \textbf{Case 1 — Target UI design.}
    }
    \label{fig:case1_ref}
\end{figure*}

\begin{figure*}[p]
    \centering
    \includegraphics[width=\textwidth]{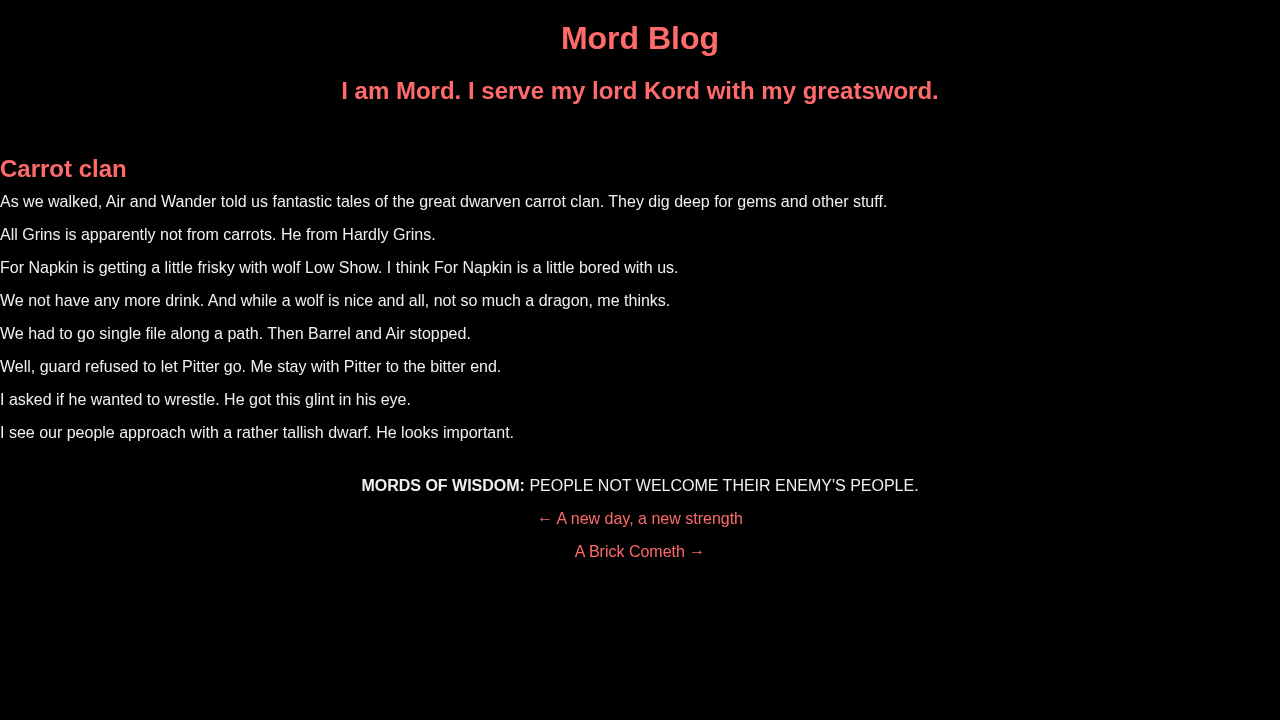}
    \caption{
        \textbf{Case 1 — Qwen2.5-VL-7B single-step output.}
    }
    \label{fig:case1_7b}
\end{figure*}

\begin{figure*}[p]
    \centering
    \includegraphics[width=\textwidth]{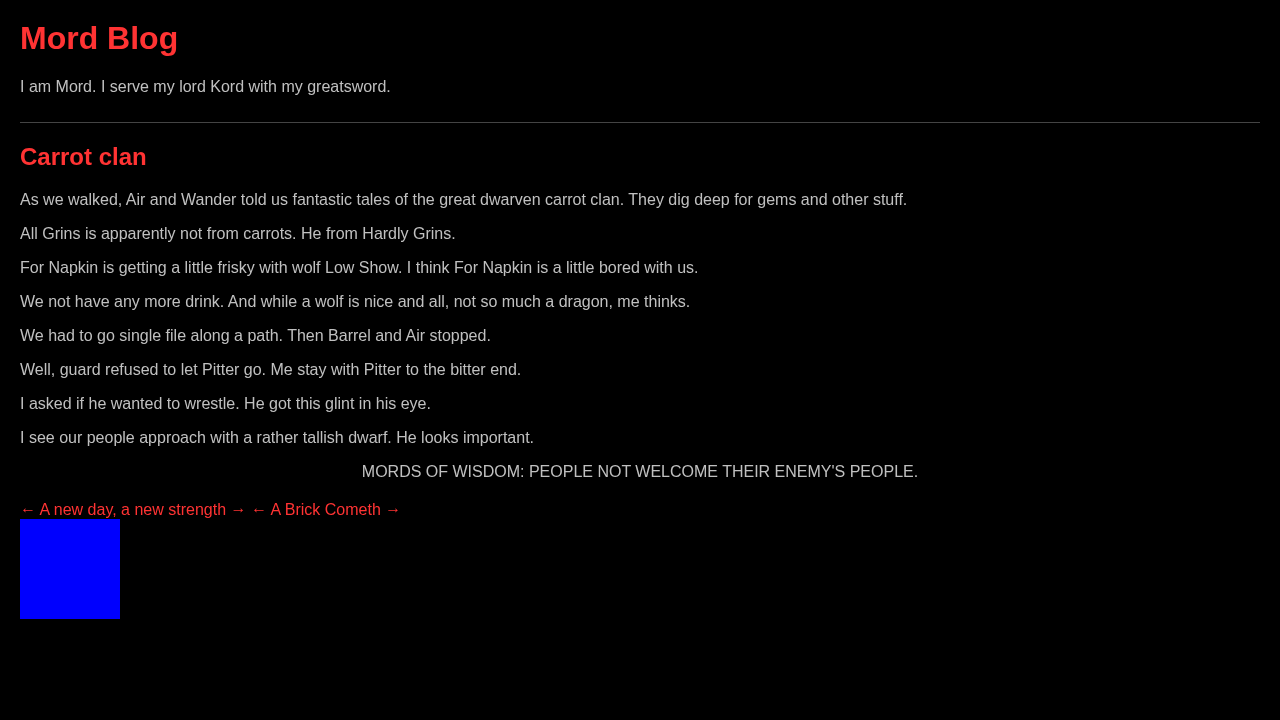}
    \caption{
        \textbf{Case 1 — Qwen2.5-VL-72B single-step output.}
    }
    \label{fig:case1_72b}
\end{figure*}

\begin{figure*}[p]
    \centering
    \includegraphics[width=\textwidth]{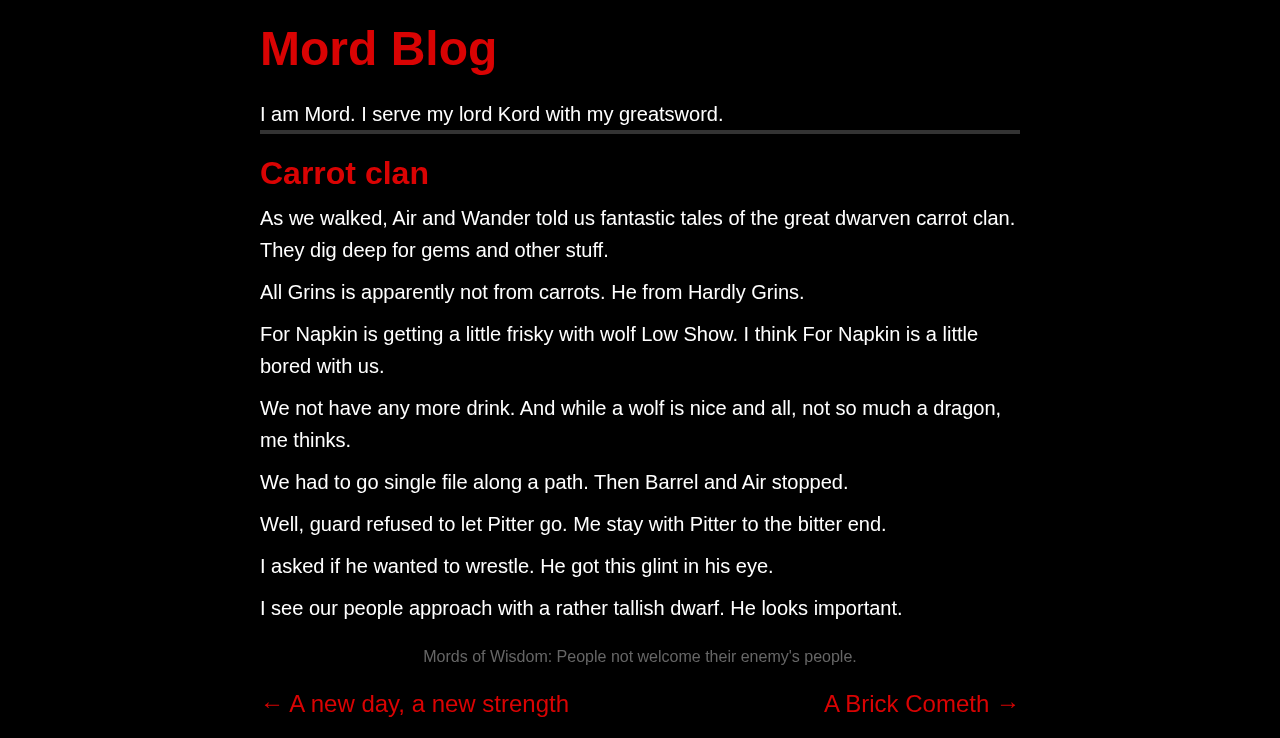}
    \caption{
        \textbf{Case 1 — VisRefiner-7B single-step output.}
    }
    \label{fig:case1_visrefiner}
\end{figure*}

\begin{figure*}[t]
    \centering
    \begin{tabular}{cccc}
        \includegraphics[width=0.24\linewidth]{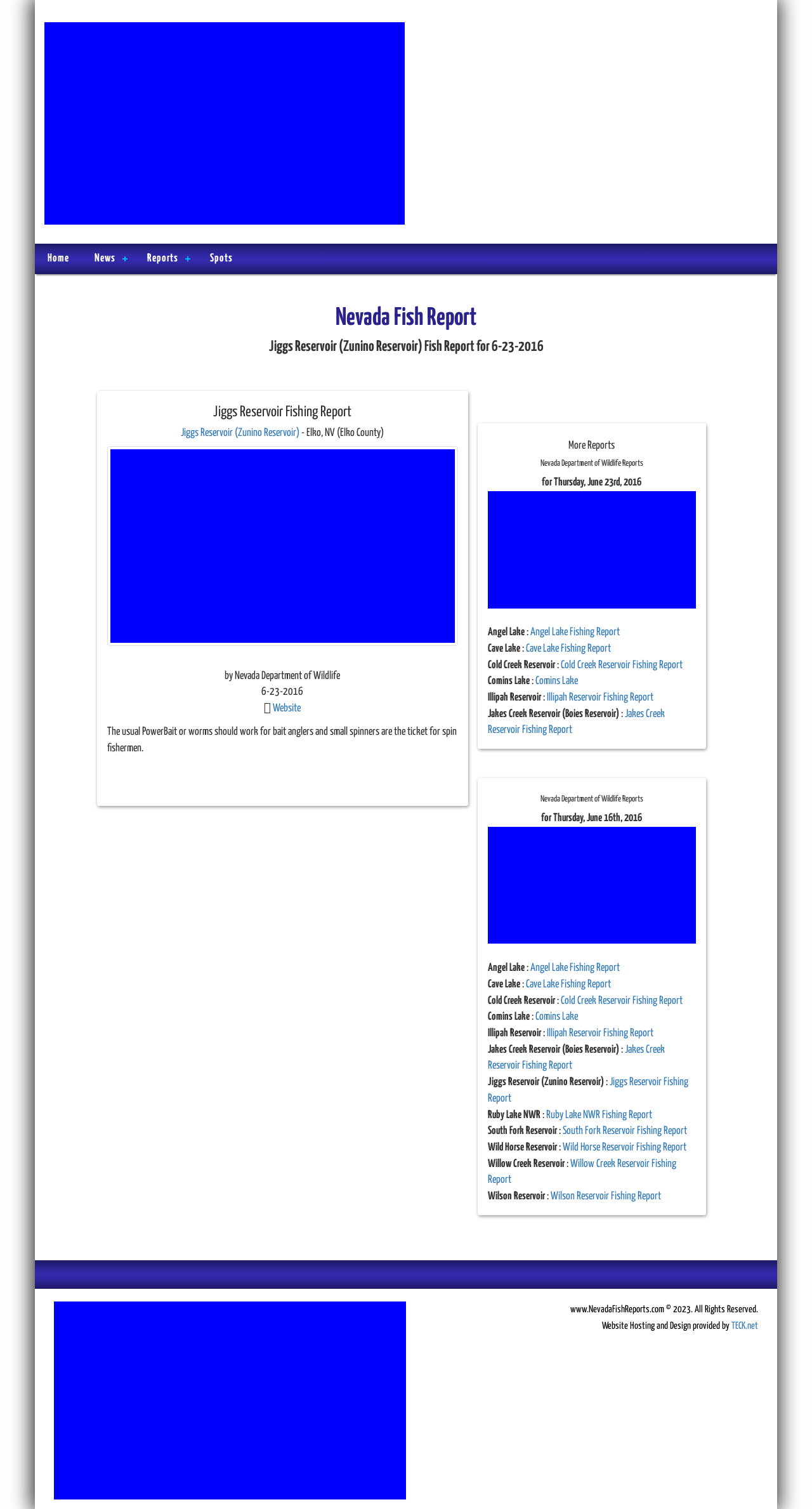} &
        \includegraphics[width=0.24\linewidth]{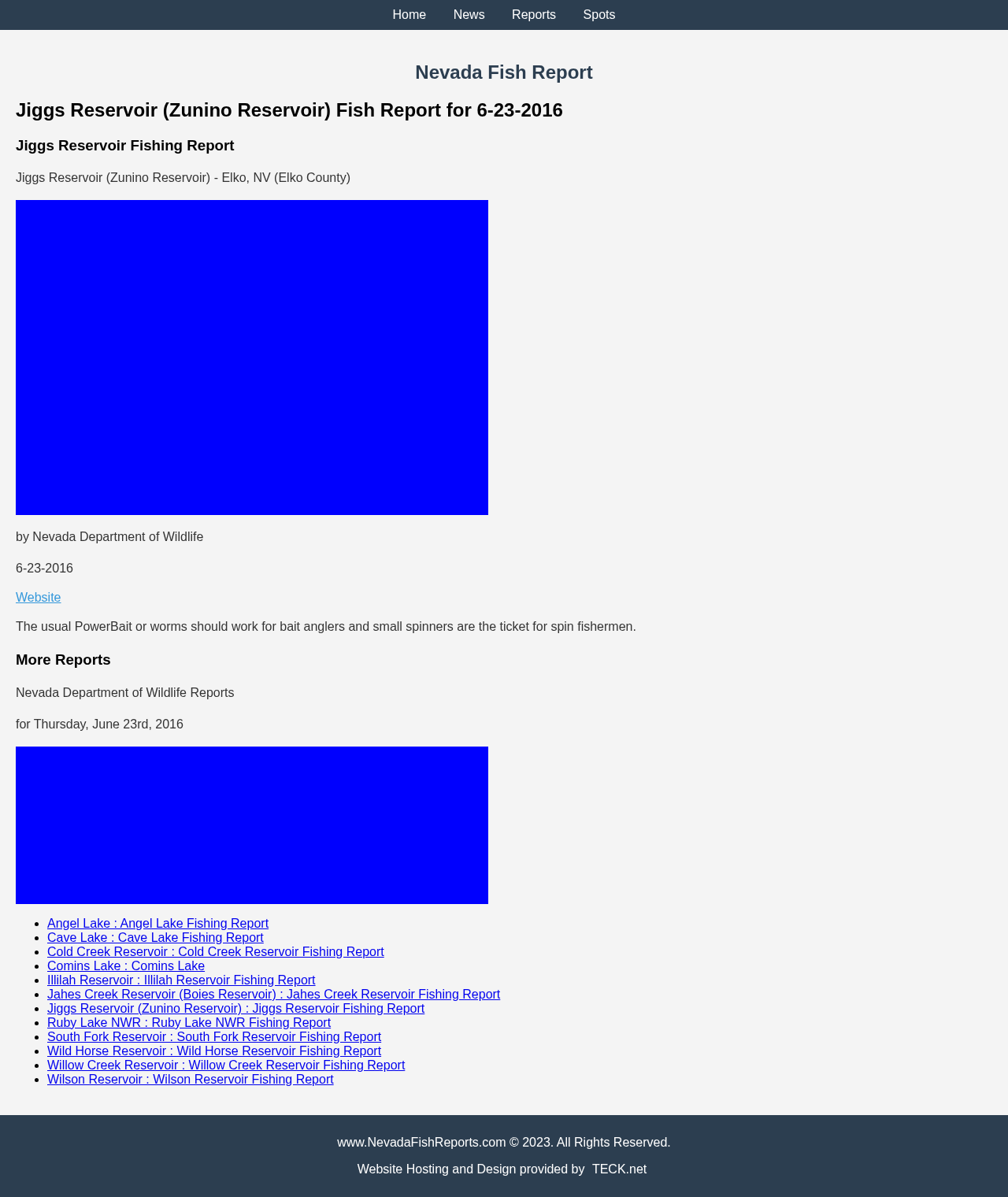} &
        \includegraphics[width=0.24\linewidth]{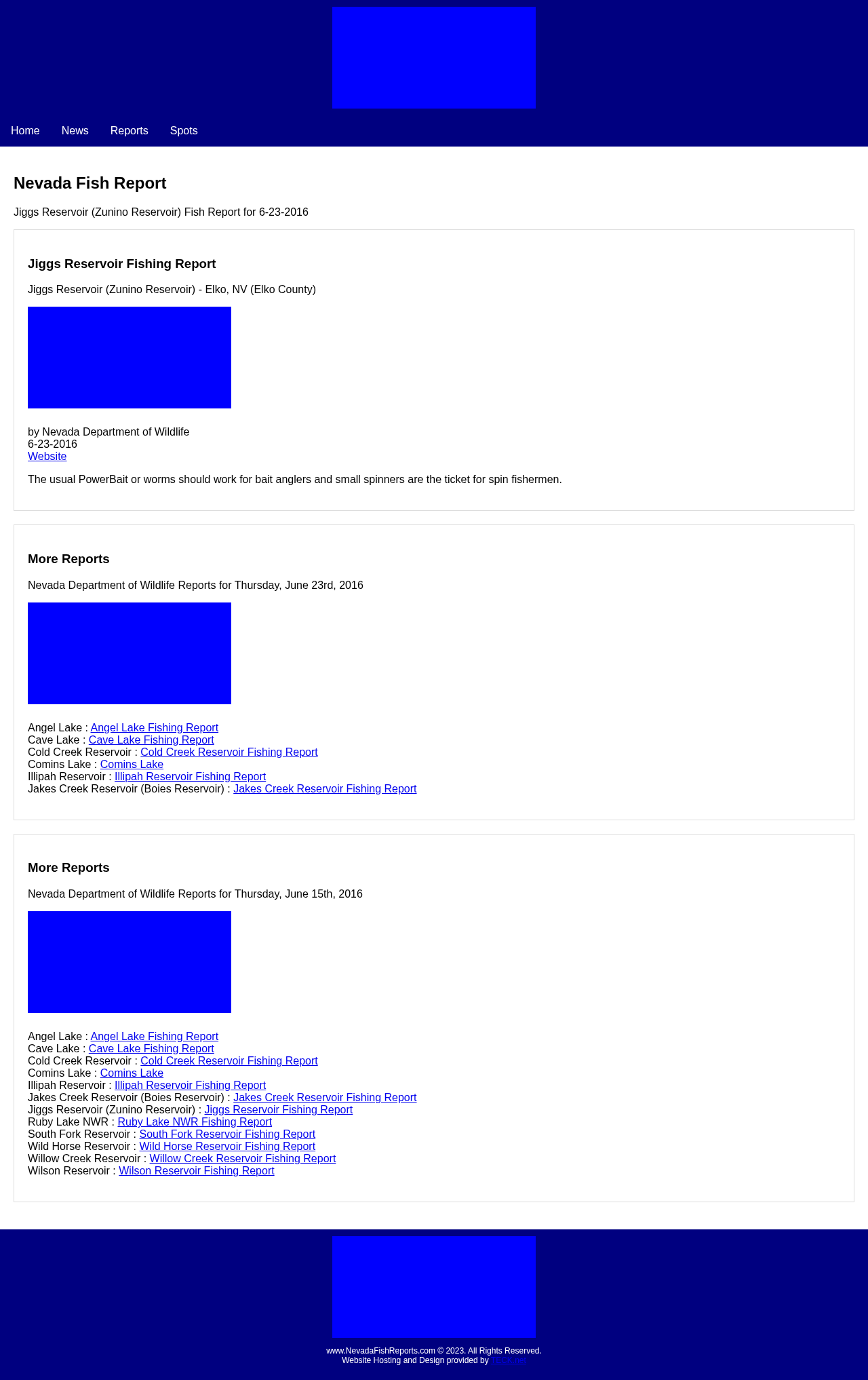} &
        \includegraphics[width=0.24\linewidth]{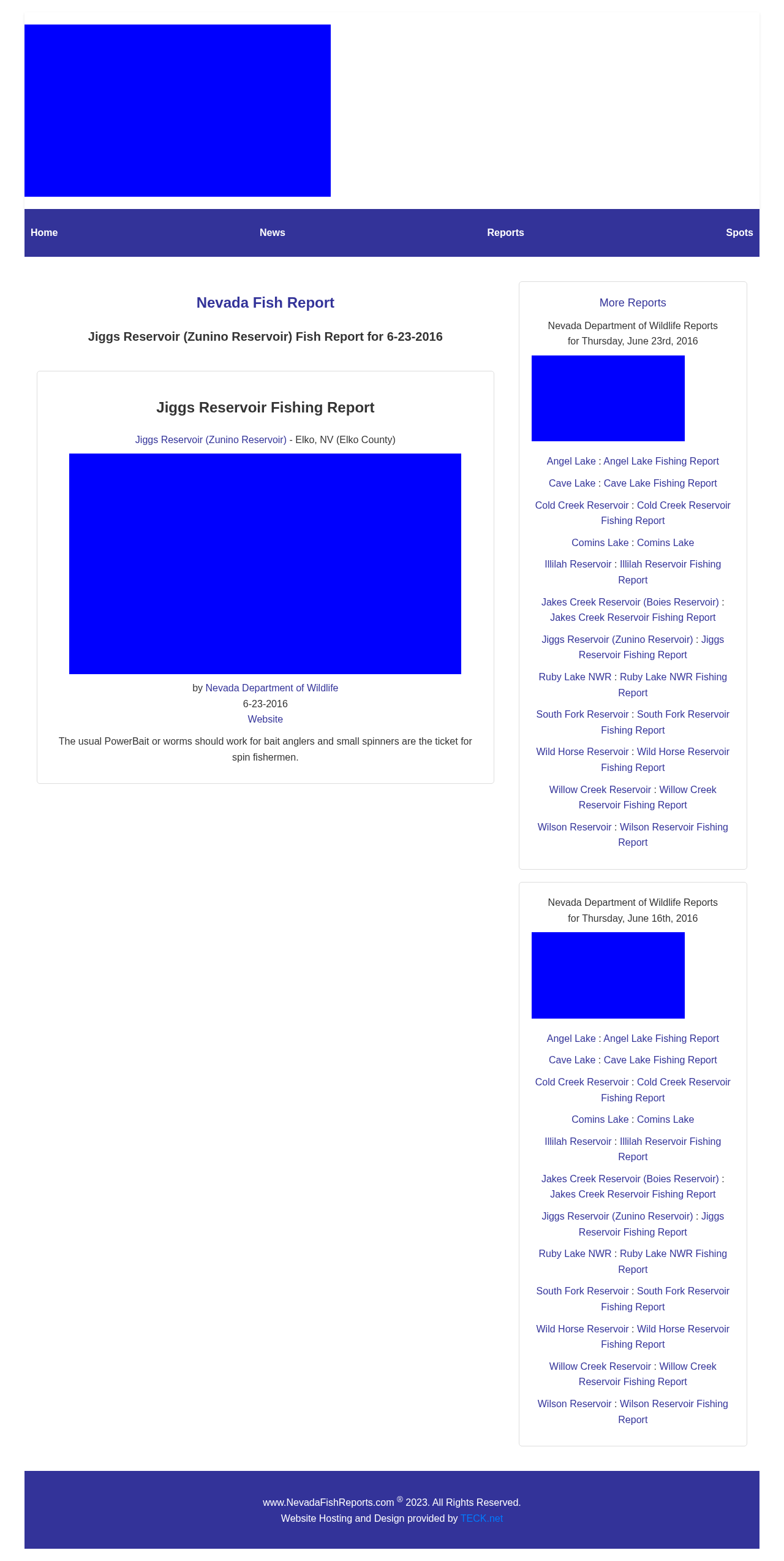} \\
        \small Target & \small Qwen2.5-VL-7B & \small Qwen2.5-VL-72B & \small VisRefiner-7B
    \end{tabular}
    \caption{
        \textbf{Case 2 — Single-step screenshot-to-code comparison.}
    }
    \label{fig:case_single_step}
\end{figure*}

We present examples to illustrate the behavior of VisRefiner in both\textbf{single-step generation} and \textbf{self-refinement}. As shown in Figures~\ref{fig:case1_ref}--\ref{fig:case_single_step}, VisRefiner produces noticeably more accurate layouts than baseline MLLMs in a single forward pass. In Figures~\ref{fig:self_ref_caseA} and \ref{fig:self_ref_caseB}, we further demonstrate its self-refinement.

\clearpage

\begin{figure*}[p]
    \centering
    \begin{tabular}{ccc}
        \includegraphics[width=0.3\linewidth]{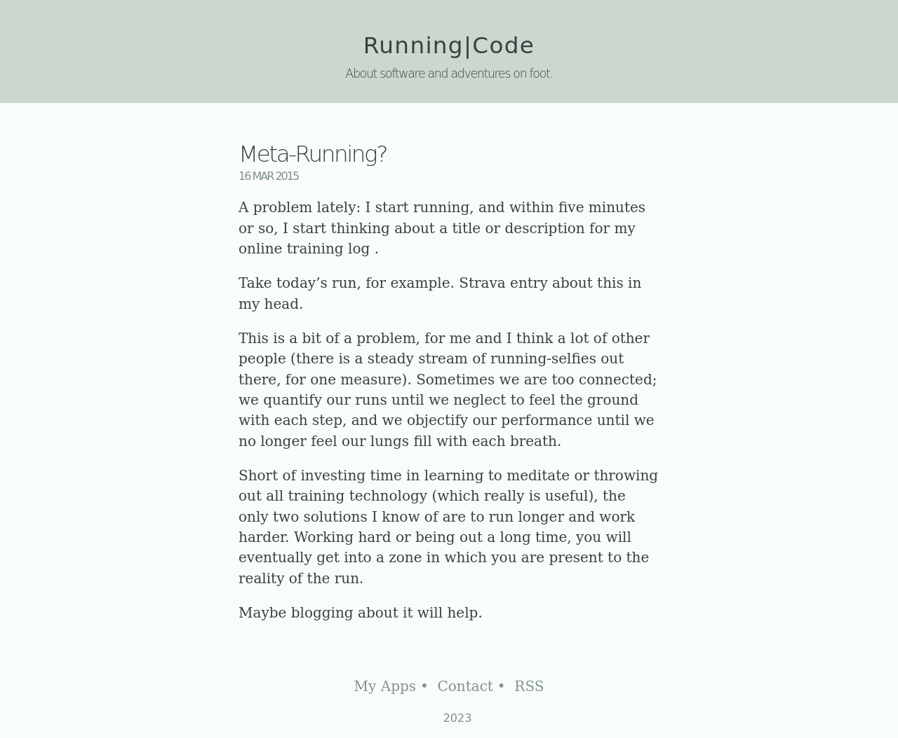} &
        \includegraphics[width=0.3\linewidth]{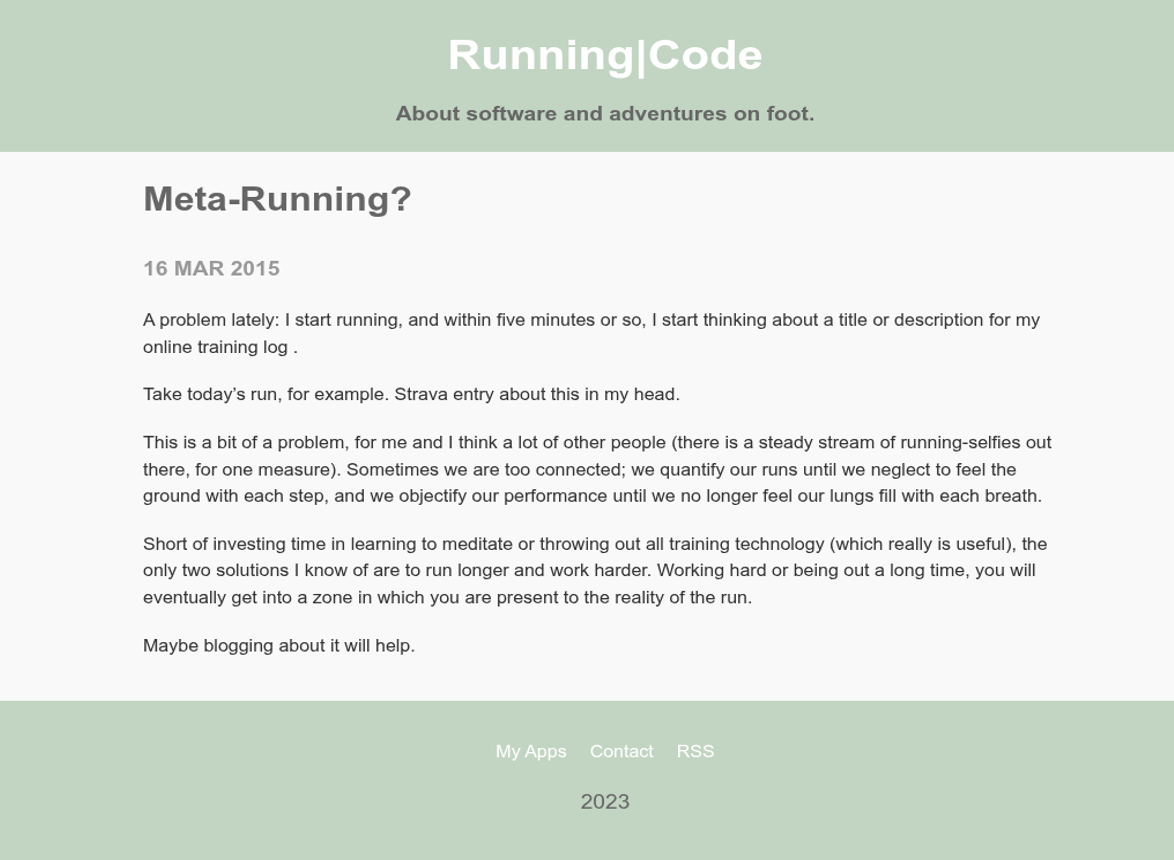} &
        \includegraphics[width=0.3\linewidth]{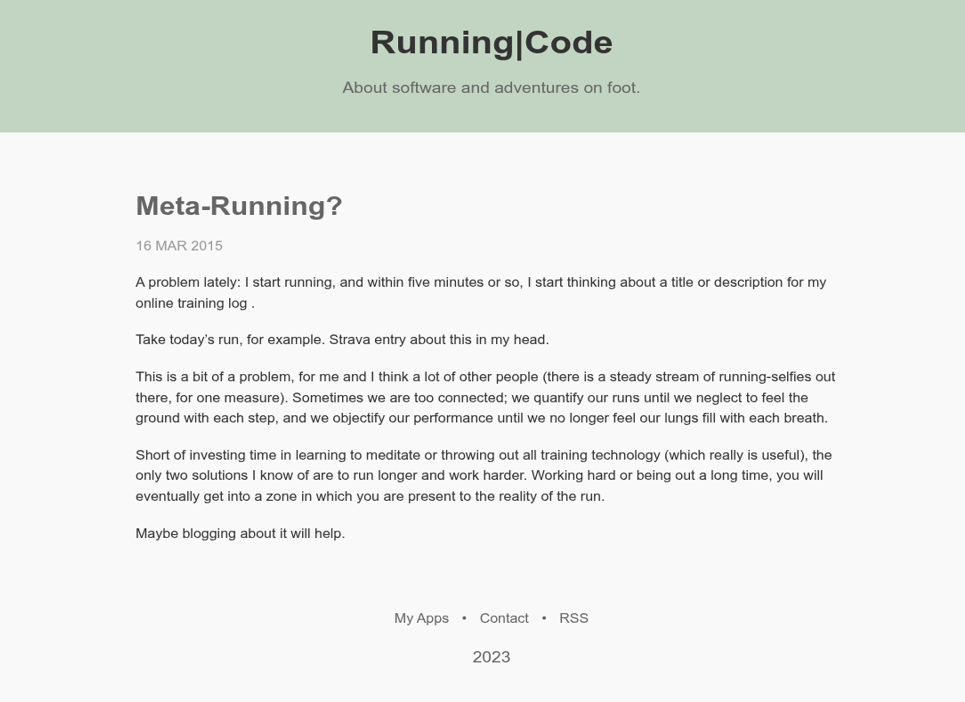} \\
        \small Target & \small Turn 0 & \small Turn 1
    \end{tabular}

    \vspace{4pt}

    \begin{tabular}{ccc}
        \includegraphics[width=0.3\linewidth]{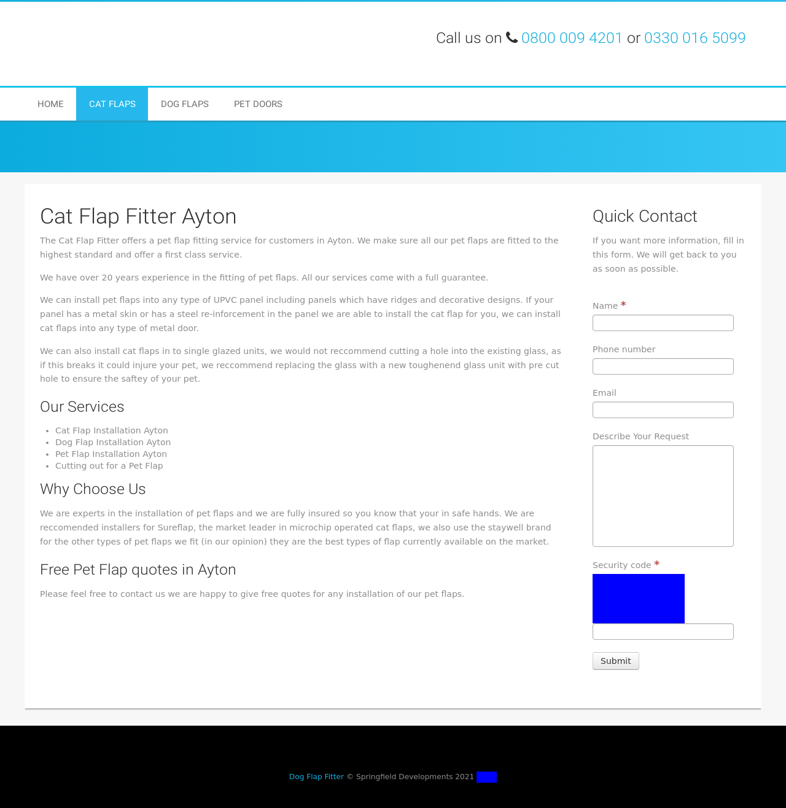} &
        \includegraphics[width=0.3\linewidth]{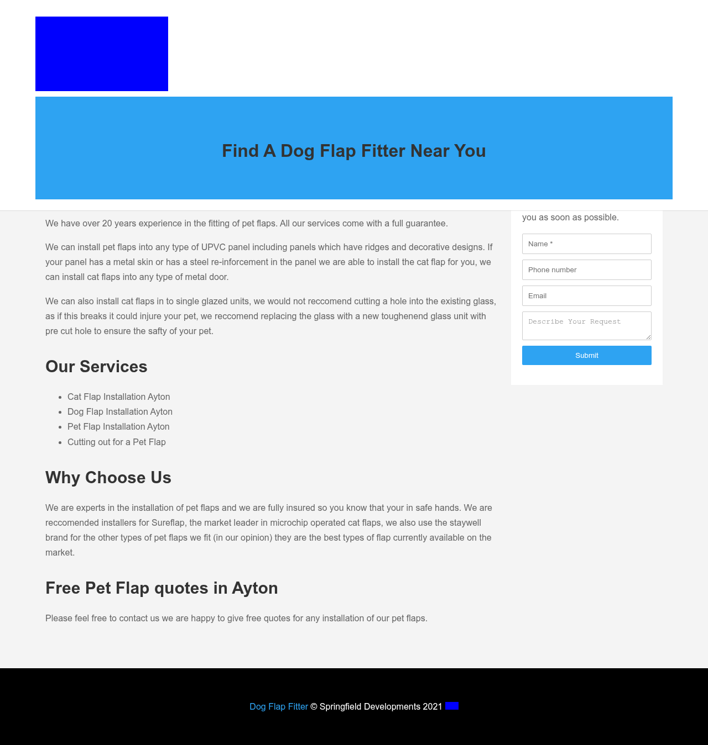} &
        \includegraphics[width=0.3\linewidth]{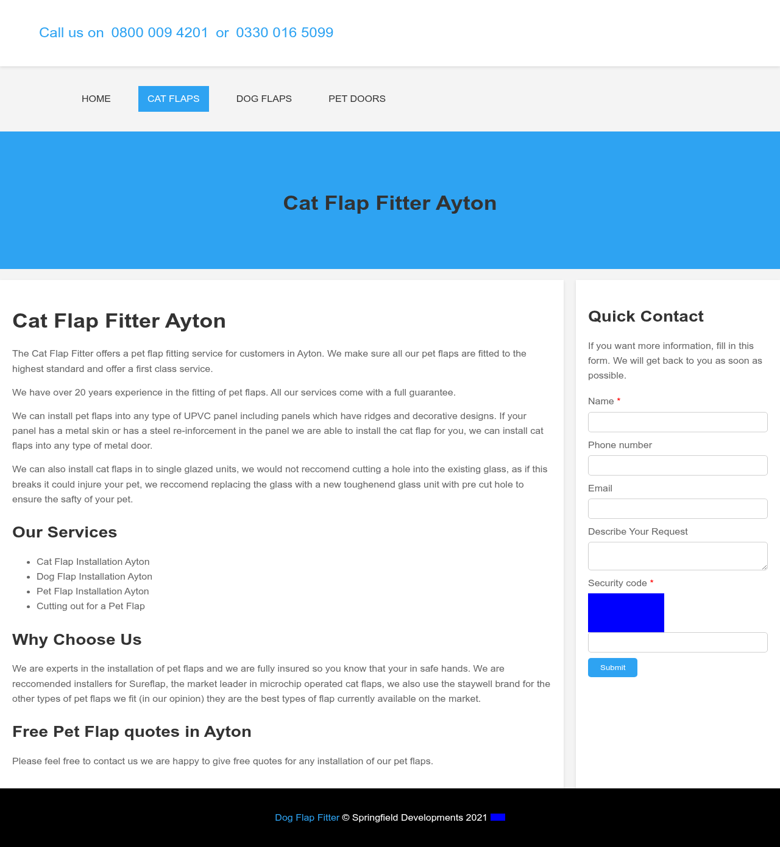} \\
        \small Target & \small Turn 0 & \small Turn 1
    \end{tabular}

    \caption{
        \textbf{Self-refinement examples.}
        \textbf{Top:} Case A. \textbf{Bottom:} Case B.
        Each case shows the target webpage and two iterative refinement turns.
    }
    \label{fig:self_ref_cases}
\end{figure*}

\end{document}